\documentclass[%
preprint,
nofootinbib,
 amsmath,amssymb,
 aps,
]{revtex4-2}

\usepackage{graphicx}
\usepackage{dcolumn}
\usepackage{bm}
\usepackage{comment}

\usepackage{xcolor}

\usepackage[normalem]{ulem}

\begin{document}

\title{Data-driven discovery of intrinsic dynamics\\
~
}

\author{Daniel Floryan}\email{dfloryan@uh.edu}

\affiliation{
  Department of Mechanical Engineering, University of Houston, Houston, TX 77204, USA
}%

\author{Michael D. Graham}\email{mdgraham@wisc.edu}

\affiliation{
  Department of Chemical and Biological Engineering, University of Wisconsin--Madison, Madison, WI 53706, USA
}%

\date{\today}

\begin{abstract}
Dynamical models underpin our ability to understand and predict the behavior of natural systems. Whether dynamical models are developed from first-principles derivations or from observational data, they are predicated on our choice of state variables. The choice of state variables is driven by convenience and intuition, and in the data-driven case the observed variables are often chosen to be the state variables. The dimensionality of these variables (and consequently the dynamical models) can be arbitrarily large, obscuring the underlying behavior of the system. In truth, these variables are often highly redundant and the system is driven by a much smaller set of latent intrinsic variables. In this study, we combine the mathematical theory of manifolds with the representational capacity of neural networks to develop a method that learns a system's intrinsic state variables directly from time series data, and also learns predictive models for their dynamics. What distinguishes our method is its ability to reduce data to the intrinsic dimensionality of the nonlinear manifold they live on. This ability is enabled by the concepts of charts and atlases from the theory of manifolds, whereby a manifold is represented by a collection of patches that are sewn together---a necessary representation to attain intrinsic dimensionality. We demonstrate this approach on several high-dimensional systems with low-dimensional behavior. The resulting framework provides the ability to develop dynamical models of the lowest possible dimension, capturing the essence of a system.
\end{abstract}

\maketitle

\section{Introduction}	
\label{sec:intro}	

Dynamical models are fundamental to our ability to model systems in engineering and the sciences. Accurate models of their dynamics enable deeper understanding of these systems, as well as the ability to predict their future behaviour. In some cases, dynamical models can be derived from first principles; the equations describing how an apple falls under the influence of gravity, for example, can be derived by applying Newton's second law. In other cases, though, no such dynamical models are available. Even when dynamical models can be derived, they may be high-dimensional to the point of obscuring the underlying behavior of a system, being difficult to analyze, and being prohibitively expensive to make predictions with. 

The latter three points have led to great efforts in developing methods that learn low-dimensional dynamical models directly from time series data \cite{watters2017visual, gonzalez2018deep, vlachas2018data, champion2019data, carlberg2019recovering, linot2020deep, maulik2020time, hasegawa2020machine, linot2021data, maulik2021reduced, rojas2021reduced, vlachas2022multiscale}. We focus on the case where rich (high-dimensional) time series data are available, which is increasingly becoming the norm in the era of big data. Even time series of a single measurement can be augmented via time delay embedding to fit this mold \cite{takens1981detecting}. Such methods typically comprise two modules: one that learns a latent representation of the state of the system, and another that learns how the latent state representation evolves forward in time \cite{watters2017visual, gonzalez2018deep, vlachas2018data, champion2019data, carlberg2019recovering, linot2020deep, maulik2020time, hasegawa2020machine, linot2021data, maulik2021reduced, rojas2021reduced, vlachas2022multiscale}. A key enabling assumption, sometimes called the manifold hypothesis \cite{fefferman2016testing}, is that the data lie on or near a low-dimensional manifold; for physical systems with dissipation, such manifolds can often be rigorously shown to exist \cite{hopf1948mathematical, foias1988inertial, temam1994estimates, doering1995applied}. These manifolds enable a low-dimensional latent state representation, and hence, low-dimensional dynamical models. Linear manifold learning techniques, such as principal component analysis, cannot learn the nonlinear manifolds that represent most systems in nature. To do so, we require nonlinear methods, some of which are developed in \cite{scholkopf1998nonlinear, tenenbaum2000global, roweis2000nonlinear, belkin2003laplacian, donoho2003hessian, van2008visualizing, goodfellow2016deep} and reviewed in \cite{ma2012manifold}. 

In this work, we present a method that learns minimal-dimensional dynamical models directly from data. What distinguishes our method is its ability, in principle, to reduce data to the intrinsic dimensionality of the nonlinear manifold they live on without any loss of information. That is, in principle our method has the ability to form latent representations of data that are minimal-dimensional and lossless. 

To obtain models of intrinsic dimensionality, we combine the rigorous mathematical theory of manifolds with the approximation capability of neural networks. Our method decomposes a manifold into overlapping patches, independently reduces each patch to the intrinsic dimensionality via a collection of autoencoders, and learns models of the dynamics on each patch using neural networks. The patches are sewn together to obtain a global dynamical model  without needing a global parameterization of the manifold. Each of these elements is crucial to obtaining an accurate minimal model, explaining why other local approaches are unable to do so \cite{bregler1994surface, hinton1995recognizing, kambhatla1997dimension, roweis2002global, brand2003charting, amsallem2012nonlinear, pitelis2013learning, schonsheck2019chart}. For example, \cite{bregler1994surface, hinton1995recognizing, kambhatla1997dimension, roweis2002global, brand2003charting} form a global parameterization of a manifold, limiting the degree of dimension reduction possible. In the language of topology, a patch and its corresponding autoencoder would be called a chart, and the collection of all the charts is called an atlas. Hence, we refer to our method as \emph{Charts and Atlases for Nonlinear Data-Driven Dynamics on Manifolds}---CANDyMan, for short.

\section{A primer on manifolds}	
\label{sec:mani}

Before describing our method, we review and motivate elements of the theory of manifolds. 

\begin{figure}
\centering
\includegraphics[width=0.5\linewidth]{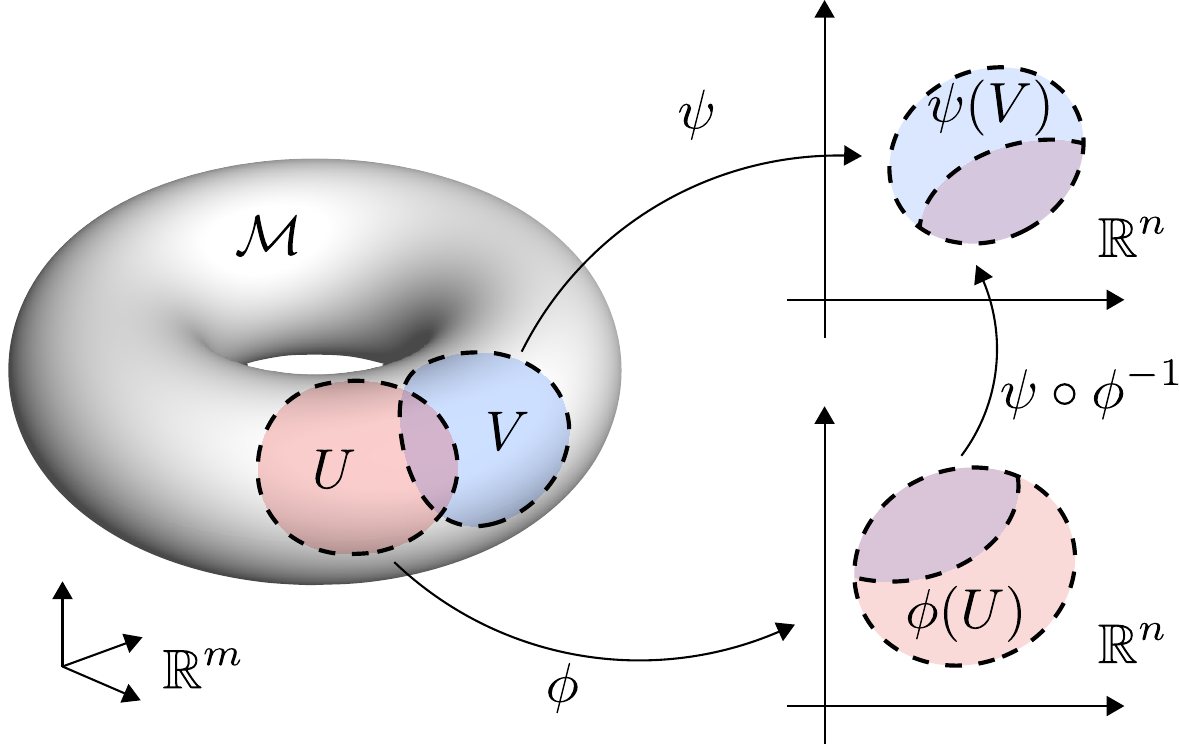}
\caption{Two charts, $(U,\phi)$ and $(V,\psi)$, of an $n$-dimensional manifold, $\mathcal{M}$, embedded in $\mathbb{R}^m$, and a transition map, $\psi\circ\phi^{-1}$.}	
\label{fig:charts}
\end{figure}

The word ``manifold'' evokes an image of a curved surface embedded in a higher-dimensional Euclidean space; this is the relevant scenario for us, in which case our manifolds are, strictly speaking, submanifolds of $\mathbb{R}^m$. A manifold $\mathcal{M}$ is a set that is locally Euclidean, meaning that every point on the manifold has an open neighbourhood that can be mapped back and forth from and to a Euclidean space $\mathbb{R}^n$; the map and its inverse are continuous, and $n \le m$ is called the dimension of the manifold, which we called the intrinsic dimensionality earlier. Informally, one may think of breaking a manifold into overlapping patches such that each patch can be flattened. This representation is sketched in figure~\ref{fig:charts}, and is formalized by the concept of charts. 

A chart is a pair $(U, \phi)$, where the coordinate domain $U$ is an open subset of the manifold, and the coordinate map $\phi$ is a homeomorphism mapping $U$ to $\phi(U) \subseteq \mathbb{R}^n$. A point $p \in U$ is mapped to $\phi(p)$, called the local coordinates on $U$. If the coordinate domain of a chart can cover the entire manifold, then the manifold is homeomorphic to $\mathbb{R}^n$. In general, however, the coordinate domain of a chart cannot cover the entire manifold, so there will be other charts, and each point on the manifold must belong to the coordinate domain of at least one chart. An atlas is a collection of charts whose coordinate domains cover the manifold, and forms a working representation of a manifold. 

To switch between local coordinates, we use transition maps. Given two charts, $(U, \phi)$ and $(V, \psi)$, whose coordinate domains intersect, the transition map from $\phi$ to $\psi$ is given by $\psi \circ \phi^{-1}: \phi(U \cap V) \rightarrow \psi(U \cap V)$; it is also a homeomorphism. We refer the interested reader to~\cite{lee2013introduction} for more details. 

\begin{figure}	
\centering
\includegraphics[width=\linewidth]{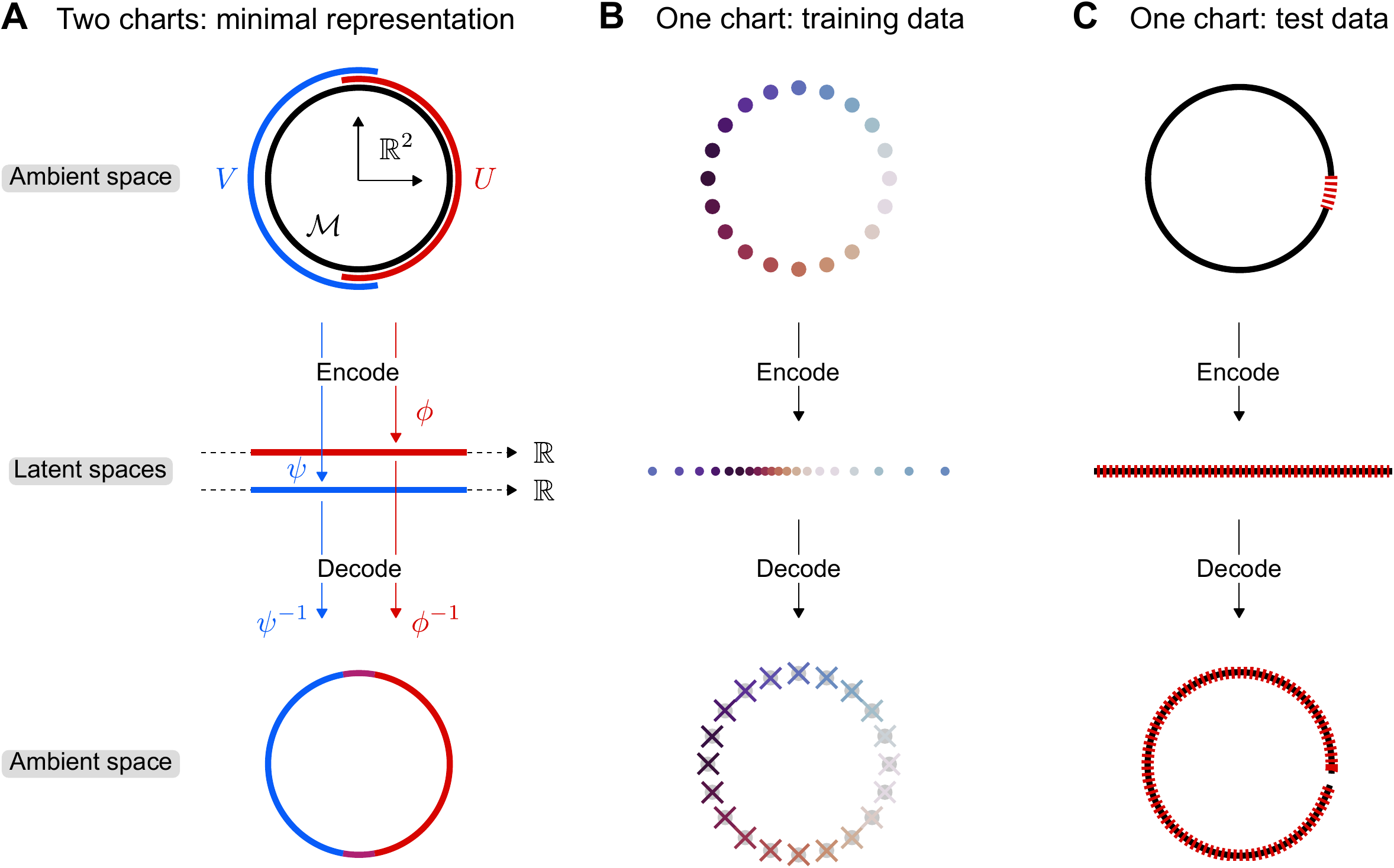}
\caption{(A) A minimal one-dimensional representation of a circle can be attained by using two charts. (B) An autoencoder (one chart) is trained to map points on a circle ($\bullet$) to one dimension and back to the circle ($\times$). (C) When the autoencoder is applied to the entire circle, a gap emerges. The gap emerges because the encoder is, and always will be, a non-one-to-one function. }
\label{fig:circleFail}
\end{figure}

A simple circle in the plane elucidates why charts are useful. A circle is a one-dimensional manifold, so we should be able to continuously map back and forth between the $(x,y)$ coordinates on a circle and $\mathbb{R}$. That is, we should be able to find a one-dimensional latent representation of the circle. But this can only be done locally, not globally, since the circle closes on itself. Any attempt to produce a global one-dimensional latent representation of the circle will result in a discontinuity or a map that is not one-to-one, as we demonstrate below. Instead, we must break the circle into at least two overlapping arcs, each of which can be mapped homeomorphically to $\mathbb{R}$. That is, we must use at least two charts, as shown in figure~\ref{fig:circleFail}A. 

Popular manifold learning methods, however, attempt to produce a global latent representation. Thus, they will be unable to reduce even the humble circle to one dimension and reconstruct the circle from the one-dimensional representation, as we demonstrate in the Supplementary Information. We show representative results in figure~\ref{fig:circleFail}B--C, for which we trained an undercomplete autoencoder, with a bottleneck dimension of one, on 20 $(x,y)$ pairs evenly spaced on the unit circle. The results are insidious: the autoencoder reconstructs the training data incredibly well from their one-dimensional encodings, seemingly able to represent the circle in a one-dimensional latent space. Upon closer inspection, however, the blue points occupy the opposite ends of the latent space, whereas they are near each other in the ambient space, hinting at a problem. When we apply the autoencoder to the entire circle\footnote{We encoded and decoded 10,000 equally spaced points covering angles from 0 to $2 \pi$.}, a gap emerges (see figure~\ref{fig:circleFail}C). To elucidate why, we have drawn part of the circle with a solid black line, and the rest with a dashed red line. Because the data are continuous and the encoder is a continuous function from $\mathbb{R}^2$ to $\mathbb{R}$, the dashed red line must connect the two ends of the solid black line in the latent space, just as it does in the ambient space. This is only possible if the dashed red line lies on top of the solid black line in the latent space. As a result, the encoder is necessarily not one-to-one, mapping different parts of the circle on top of each other. It is then impossible to recover the original circle from the latent space since the solid black and dashed red portions cannot be distinguished; the solid black and dashed red portions must coincide after the decoding step, resulting in a gap. This is insidious because we can attain arbitrarily small training error but will never attain generalization. This problem is due to the geometry of the circle; it is independent of the amount of training data, network architecture, etc., and is emblematic of the limitations of global manifold learning methods. 


It is worth pointing out that methods that attempt to produce a global coordinate system, which we may think of as single-chart methods, are still able to achieve some degree of dimension reduction without loss of information, though generally not down to the intrinsic dimension. The strong Whitney embedding theorem states that any smooth real $n$-dimensional manifold can be smoothly embedded in $\mathbb{R}^{2n}$ \cite{whitney1944self, lee2013introduction}. This means that a single-chart method can theoretically reduce the dimension to $2n$ in the worst case, compared to $n$ for a multi-chart method. For example, a single-chart autoencoder would be able to reduce a circle in three-dimensional $(x,y,z)$ space to two dimensions, but not to its intrinsic dimension of one. Despite the theoretical guarantee of the strong Whitney embedding theorem, in Section~\ref{sec:ksburst} we will show an example that suggests that the practical gap between multi-chart and single-chart methods is even greater than the theoretical gap. 

We note that a limited set of work has recognized the concepts of charts and atlases~\cite{brand2003charting, pitelis2013learning, schonsheck2019chart}. Brand \cite{brand2003charting} was the first to explicitly mention charts, but uses them to form a global parameterization of a manifold, leading to the limitations of global manifold learning methods. More recent methods follow the formalism of charts and atlases more closely~\cite{pitelis2013learning, schonsheck2019chart}, but would falter in the context of dynamics: Pitelis \textit{et al.} {\cite{pitelis2013learning} use linear approximations for the coordinate maps, which would lead to discontinuities when transitioning between charts; similarly, the method of  Schonsheck \textit{et al.} \cite{schonsheck2019chart} would also lead to non-smooth dynamics.

\section{Learning an atlas and dynamics on a manifold}
\label{sec:method}

With the utility of an atlas of multiple charts now clear, we describe our method in detail. Consider a discrete dynamical system of the form
\begin{equation}
  \label{eq:method1}
  x_{i+1} = F(x_i).
\end{equation}
This encompasses continuous dynamical systems of the form
\begin{equation}
  \label{eq:method2}
  \frac{\text{d}}{\text{d}t} x = f(x)
\end{equation}
since they may be written as 
\begin{equation}
  \label{eq:method3}
  x(t + \Delta t) = F(x(t)) = x(t) + \int_t^{t + \Delta t} f(x(\tau)) \, \text{d}\tau. 
\end{equation}
The state of the system $x \in \mathbb{R}^m$ evolves with time, with the dynamics given by $F$. 

Suppose we have a dataset generated by our dynamical system. Namely, we have a set of pairs of vectors, $\{(x_i, x_i')\}_{i=1}^N$, with $x_i, x_i' \in \mathbb{R}^m$ and $x_i' = F(x_i)$ for $i = 1, \ldots, N$. For a continuous dynamical system, we assume that $x_i$ and $x_i'$ are sampled from the continuous dynamical system at times $\Delta t$ apart for $i = 1, \ldots, N$. Usually, such a dataset comes from a single time series $\{x_i\}_{i=1}^{N+1}$, and $x_i' = x_{i+1}$ for $i = 1, \ldots, N$.	

It is often the case that the dynamics, and hence the data, live on a submanifold $\mathcal{M} \subset \mathbb{R}^m$ of dimension $n \ll m$. Dissipative systems constitute an important case, where at long times, the dynamics approach an invariant manifold. Even for systems described by partial differential equations, which are formally infinite-dimensional, the presence of dissipation can lead to the long-time dynamics residing on a finite-dimensional submanifold \cite{hopf1948mathematical, foias1988inertial, temam1994estimates, doering1995applied}. Our goal is to learn these manifolds, and the dynamics on them, directly from the data. 

The method consists of two steps: first, we learn an atlas of charts; second, we learn the dynamics in the local coordinates of each chart. 
The key step is to patch the local descriptions together to form a global one. We describe the method below, simultaneously demonstrating each step on the simple example of a particle moving counterclockwise around the unit circle at constant speed. In this example, the data are embedded in $\mathbb{R}^2$, but live on the one-dimensional manifold $S^1$. In Section~\ref{sec:ex}, we demonstrate the method on more complex examples. 


\subsection{Learning an atlas}
\label{sec:atlas}	

First, we must decompose the data manifold into patches; that is, we must learn a set of coordinate domains. To start, we use $k$-means clustering \cite{macqueen1967some, steinhaus1957division, lloyd1982least, forgy1965cluster, pedregosa2011scikit} to partition the data into disjoint sets. We use $k$-means clustering because it provides a rational way to partition a dataset and offers algorithmic benefits that will be important later. These sets do not yet form the coordinate domains of charts because they do not overlap. To make the sets overlap, we start by building a graph from the data, placing undirected edges between a data point and its $K$ nearest neighbours, for every data point. We then expand each cluster along the graph, giving overlapping coordinate domains and causing some data points to be members of multiple clusters (figure~\ref{fig:method}A; in this case, each cluster was expanded by two neighbours along the graph). 

With the coordinate domains in hand, we next learn the coordinate maps and their inverses. The universal approximation theorem guarantees that all the coordinate maps, as well as their inverses, can be approximated arbitrarily well by sufficiently large neural networks\footnote{In addition to approximation error, estimation error and optimization error are important considerations~\cite{bottou2007tradeoffs}.} \cite{cybenko1989approximation, hornik1991approximation, pinkus1999approximation}. Taken together, a neural coordinate map and its approximate inverse form an autoencoder. For each cluster, we use that cluster's data to train a deep fully-connected feedforward autoencoder that reduces the dimension of the data (figure~\ref{fig:method}B-C). In general, we do not know the dimension of the underlying manifold, so we estimate it by monitoring the reconstruction error of the autoencoder as a function of the latent dimension (cf.~\cite{linot2020deep, vlachas2022multiscale}; alternately, one can monitor the rank of the data covariance in the latent space~\cite{jing2020implicit}). The resulting encoder is the coordinate map of the chart, and the decoder is its approximate inverse. Models besides neural networks can be used; for example, principal component analysis can be used to project the data to a lower dimension, and kernel ridge regression used to construct the approximate inverse of that projection. An approximate transition map between charts is formed by composing the encoder of one chart with the decoder of the other chart. 

\begin{figure}
\centering
\includegraphics[width=0.8\linewidth]{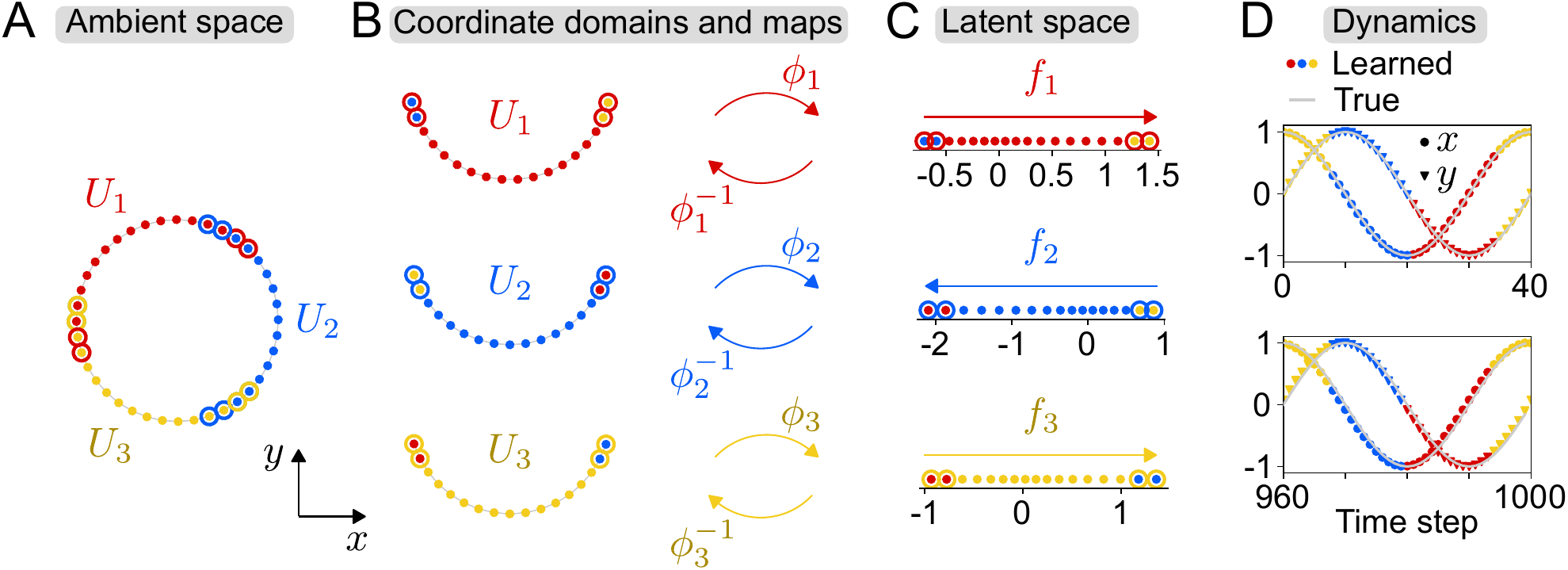}
\caption{The method consists of three basic steps: (A) learning coordinate domains ($U_1$, $U_2$, and $U_3$ here), (B) learning coordinate maps and their inverses, and (C) learning the dynamics in the local coordinates. In D, we show that the first (top) and 25th (bottom) cycles of motion are well predicted. Filled markers are interior points, and large open markers are border points. }
\label{fig:method}
\end{figure}

Note that the ranges of values of the local coordinates differ amongst the three sets of local coordinates in figure~\ref{fig:method}, and that the densities of the data are not uniform in the local coordinates as they are on the unit circle. The definition of a manifold places no restrictions on the values of the local coordinates; any local coordinates will do as long as the coordinate map is homeomorphic. So, we have not restricted our autoencoders, e.g., by regularizing them in some way. As a result, the sets of local coordinates differ from each other and the distributions of data are distorted by the coordinate maps. In a later example, we find that normalizing the local coordinates is helpful when learning dynamics.

\subsection{Learning the dynamics}
\label{sec:dyn}

The dynamics on the manifold is given by a map $F_\mathcal{M}: \mathcal{M} \rightarrow \mathcal{M}$. In each chart $\alpha$'s local coordinates, the dynamics are given by $f_\alpha = \phi_\alpha \circ F_\mathcal{M}  \circ \phi_\alpha^{-1} : \mathbb{R}^n \rightarrow \mathbb{R}^n$. That is, for the chart $(U_\alpha, \phi_\alpha)$ and a point $x \in U_\alpha$, the local coordinates of $x$, $\phi_\alpha(x)$, are mapped to $f_\alpha(\phi_\alpha(x))$ under the dynamics. The map $f_\alpha$ can be approximated arbitrarily well by a sufficiently large neural network\footnote{In addition to approximation error, estimation error and optimization error are important considerations~\cite{bottou2007tradeoffs}.} \cite{cybenko1989approximation, hornik1991approximation, pinkus1999approximation}, and it gives the low-dimensional dynamics directly (figure~\ref{fig:method}C). We train the neural network to minimize the loss 
\begin{equation}
  \label{eq:dyn1}
  \mathcal{L} = \frac{1}{| \mathcal{I}_\alpha |} \sum_{i \in \mathcal{I}_\alpha} \left\Vert f_\alpha(\phi_\alpha(x_i)) - \phi_\alpha(x_i') \right\Vert_2^2,
\end{equation}
where $\mathcal{I}_\alpha$ is an index set tracking which training data belong to chart $\alpha$. 

To form a global picture, we need a way to transition between charts under the dynamics. For this purpose, we define the notions of interior and border data points, similar to \cite{pitelis2013learning}. We say that a data point is an interior point of a cluster if it was originally assigned to that cluster (filled markers in figure~\ref{fig:method}A), and we say that a data point is a border point of a cluster if it was assigned to that cluster during the expansion process (open markers in figure~\ref{fig:method}A). A data point may be a border point of many clusters (or even no clusters), but a data point has a unique interior assignment (as long as the original clustering of data assigned points to unique clusters, as $k$-means does). 	

Global dynamics on the manifold works as follows. Starting with an initial condition $x \in \mathcal{M}$, we must find which chart's interior it is in. This is simply the chart $\alpha$ whose corresponding cluster centroid $x_\alpha$ is closest to $x$. We use chart $\alpha$'s coordinate map to map $x$ to that chart's local coordinates, $\phi_\alpha(x)$. This initial step is the only one where calculations are performed in the ambient space $\mathbb{R}^m$, and the calculations only involve the cluster centroids. Next, we apply the dynamics that we learned in the local coordinates to map the state forward to $f_\alpha(\phi_\alpha(x))$. Each time we map the state forward, we find the closest training data point $\phi_\alpha(x_k)$ (contained in $\phi_\alpha(U_\alpha)$) in the local coordinates. If $x_k$ is an interior point of the chart, we continue to map the state forward under $f_\alpha$. If $x_k$ is a border point of the chart, instead being uniquely an interior point of chart $\beta$, we transition the state to the local coordinates of chart $\beta$ using the transition map $\phi_\beta \circ \phi_\alpha^{-1}$, and then proceed similarly, now under the dynamics $f_\beta$. For example, supposing that our state evolved under chart $\alpha$'s dynamics for $l$ time steps before transitioning to chart $\beta$, its local coordinates would be $\phi_\beta(\phi_\alpha^{-1}(f_\alpha^l(\phi_\alpha(x))))$. This process is shown in figure~\ref{fig:method}D for 25 cycles; changes in colour show chart transitions, which appear seamless.

\subsection{Smooth transitions}	
\label{sec:trans}

For the learned dynamics to be accurate, we require smooth chart transitions. Let $x$ be a point in the intersection of the coordinate domains of charts $\alpha$ and $\beta \neq \alpha$. Let $\phi_\alpha$ and $\phi_\beta$ denote the coordinate maps that we learn, and let $\phi_\alpha^{inv}$ and $\phi_\beta^{inv}$ denote their approximate inverses that we learn. In order for our model to have smooth dynamics across transitions between charts, we need $\phi_\alpha^{inv}(\phi_\alpha(x))$ and $\phi_\beta^{inv}(\phi_\beta(x))$ to be (nearly) equal for all such $x$. These are simply the reconstructions of $x$ by the two charts from $x$'s local coordinates. If they differ significantly, then there will be a discontinuity in the dynamics across transitions. Mathematically, we want to minimize
\begin{equation}
  \label{eq:trans1}
  \lVert \phi_\alpha^{inv}(\phi_\alpha(x)) - \phi_\beta^{inv}(\phi_\beta(x)) \rVert
  = \lVert [ \phi_\alpha^{inv}(\phi_\alpha(x)) - x ] - [ \phi_\beta^{inv}(\phi_\beta(x)) - x ] \rVert.
\end{equation}
The first term in brackets is chart $\alpha$'s reconstruction error, and the second term is chart $\beta$'s reconstruction error. By the triangle inequality, 
\begin{equation}
  \label{eq:trans2}
  \lVert [ \phi_\alpha^{inv}(\phi_\alpha(x)) - x ] - [ \phi_\beta^{inv}(\phi_\beta(x)) - x ] \rVert
  \le \lVert \phi_\alpha^{inv}(\phi_\alpha(x)) - x \rVert + \lVert \phi_\beta^{inv}(\phi_\beta(x)) - x \rVert. 
\end{equation}
The difference between the two charts' reconstructions of $x$ is upper-bounded by the sum of each chart's reconstruction error. We can indirectly minimize the difference in the reconstructions by minimizing each chart's reconstruction error. This argument generalizes to multiple points and more than two charts. 

By using the triangle inequality, we have decoupled the two charts. We can indirectly minimize the difference in the reconstructions while training our autoencoders separately simply by weighting training data that belong to multiple charts more heavily in each autoencoder's reconstruction loss function. Decoupling the charts allows for trivial parallelization of the training process. 

For the results shown in this work, we have not weighted data that belong to multiple charts' coordinate domains more heavily in the autoencoder reconstruction losses. We experimented with doing so, but did not see appreciable effects until the weights were so large that the autoencoders' reconstructions of data belonging to only a single coordinate domain were poor. Nevertheless, this may be an important consideration in future work.

\subsection{Time complexity}	
\label{sec:time}	
The time complexity to map a state forward one time step or to transition from one set of local coordinates to another is the time complexity of a forward pass of the associated neural network(s). This depends on the architectures of the neural networks, but a forward pass of a neural network is typically fast and would be necessary for a traditional single-chart model as well. The major expense comes from searching for a nearest neighbour. Using a $k$-d tree, the time complexity of a nearest neighbour search is $O(\log_2(N))$ for a constant dimension of the search space \cite{friedman1977algorithm}. However, the performance degrades rapidly with the dimension of the search space due to the curse of dimensionality, approaching the $O(dN)$ time complexity of brute search in a $d$-dimensional space once $d \gtrsim 10$ \cite{weber1998quantitative}. Fortunately, we only perform a single nearest neighbour search in the high-dimensional ambient space (when assigning the initial condition to a chart), and all subsequent nearest neighbour searches are performed in the local coordinates of the charts. Moreover, if using $k$-means to construct the coordinate domains, we need only search over the cluster centroids when assigning the initial condition to a chart, drastically reducing the cost. More efficient data structures, such as the ball tree, can be used if the dimensionality is still high \cite{omohundro1989five, liu2006new}.

\section{Examples}
\label{sec:ex}

We demonstrate our method on several examples of increasing complexity, from the simple periodic orbit in the plane that we already showed, to complex bursting dynamics of the Kuramoto-Sivashinsky equation. Additional examples and details about the datasets, neural network architectures, training procedures, and hyperparameters are in the Supplementary Information.

\subsection{Quasiperiodic dynamics on a torus}
\label{sec:torqp}

Building on the example of a particle moving around a circle, we now consider a particle moving along the surface of a torus. The particle travels at constant speeds in the poloidal and toroidal directions, $\sqrt{3}$ times as fast in the poloidal direction as in the toroidal direction. Since the speeds are incommensurate, the particle's orbit is quasiperiodic, densely filling the surface of the torus in the limit of infinite time. The data are the $(x, y, z)$ positions of the particle, which live on a two-dimensional submanifold of $\mathbb{R}^3$, shown in figure~\ref{fig:torus2d}A. Without using the concepts of charts and atlases, we would be unable to reduce the dimension of the system to two; we will demonstrate that our atlas-of-charts-based method successfully reduces the dimension of the system to two. 

\begin{figure}
\centering
\includegraphics[width=0.99\linewidth]{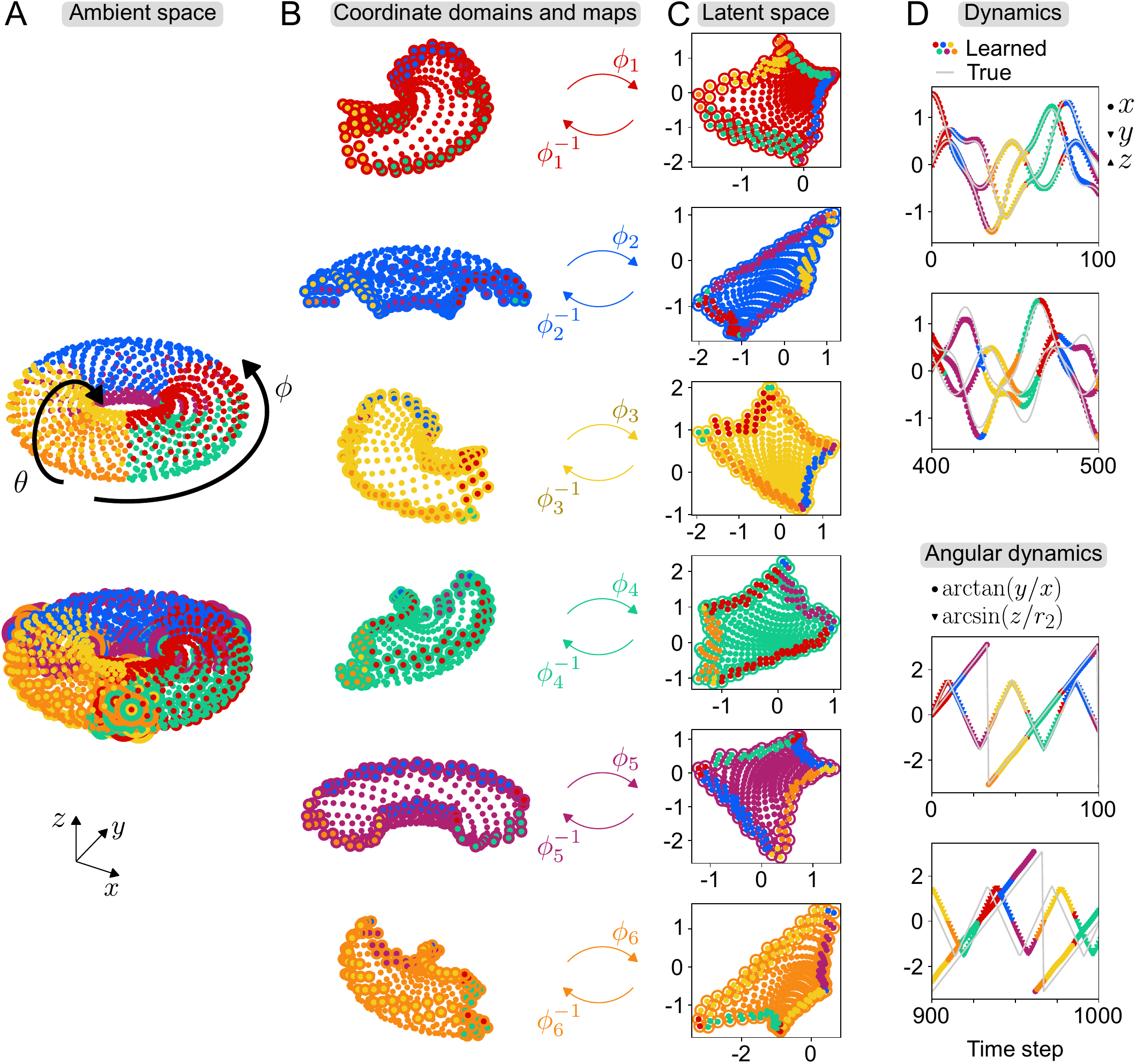}
\caption{Analogous to figure~\ref{fig:method}, but for a quasiperiodic orbit on the surface of a torus. In A, we show the data before and after coordinate domains are made to overlap; data belong to up to four coordinate domains. In D, we show that the dynamics are well predicted. The top two plots show the predicted Cartesian coordinates on the surface of the torus, and the bottom two plots show the predicted poloidal and toroidal coordinates. In each pair, the top plot shows results at short times and the bottom at much longer time, where a small amount of phase drift can be observed.}
\label{fig:torus2d}
\end{figure}

We place the data into six clusters and follow the procedure described in Section~\ref{sec:method} to create an atlas of six charts and a local two-dimensional dynamical model for each chart (figure~\ref{fig:torus2d}A--C). The choice of six charts is arbitrary; as few as three charts will work in this case. Note that some of the data points are assigned to up to four charts. Further discussion on choosing the number of charts appears in Section \ref{sec:conc}.

In figure~\ref{fig:torus2d}D, we show the trajectory that results from evolving an initial condition forward under our learned dynamical model. The learned dynamics are generally correct, with small errors in the phase speeds (roughly 0.5\% too slow in the poloidal direction and 0.5\% too fast in the toroidal direction). The transitions between charts are apparently quite smooth, even when a chart is visited for only one time step (around time step 55). When we analyze first and higher order differences between consecutive points produced by our learned dynamical model, hiccups become apparent when transitioning between charts. This is to be expected since the decoders only approximate the encoders' inverses, as discussed in Section~\ref{sec:trans}. 

For the learned model whose results are shown in figure~\ref{fig:torus2d}, we are confident that it indeed has quasiperiodic dynamics (we used the model to produce a trajectory that is 50,000 time steps long, finding no repeating pattern). However, the majority of the time we learn models that have periodic dynamics, with the period being on the order of the number of training data but varying fairly significantly (ten learned models with periodic dynamics had periods between 418 and 1892 time steps). We attribute this to the existence---in the space of weights for the neural networks for the dynamics---of Arnold tongues, leading to a mode locking phenomenon that often causes our learned model to have periodic dynamics. In short, if the dynamics of a nominally quasiperiodic system are perturbed, then the dynamics will be periodic in a finite region of parameter values; see \cite{wiggins2003introduction} for more details.

\subsection{Reaction-diffusion system}
\label{sec:rdif}

Systems in engineering and the sciences are often governed by partial differential equations (PDEs), which can generate very-high-dimensional data. To test our method on data generated by a PDE, we consider a lambda-omega reaction-diffusion system governed by 
\begin{equation}
\begin{aligned}
  \label{eq:rd}
  u_t &= [1 - (u^2 + v^2)]u + \beta (u^2 + v^2)v + d_1 (u_{xx} + u_{yy}), \\
  v_t &= -\beta(u^2 + v^2)u + [1 - (u^2 + v^2)] v + d_2 (v_{xx} + v_{yy}),
\end{aligned}
\end{equation}
with $d_1 = d_2 = 0.1$ and $\beta = 1$ \cite{champion2019data}. This system generates a spiral wave that is an attracting limit cycle in state space. The PDE is discretized on a $101 \times 101$ grid, so the data live on a one-dimensional submanifold of $\mathbb{R}^{20402}$. Without multiple charts, we would be unable to reduce the dimension of the system to one. (In the Supplementary Information, we compare our multi-chart method to the recently proposed approach of \cite{chen2021discovering}, which also attempts to learn dynamical models of intrinsic dimension from data. Theirs is a single-chart method.) A snapshot of the $u$ field is shown in figure~\ref{fig:reactDif}A, and the training data consist of 10 time units (a bit over one period of the limit cycle). 

\begin{figure}
\centering
\includegraphics[width=0.7\linewidth]{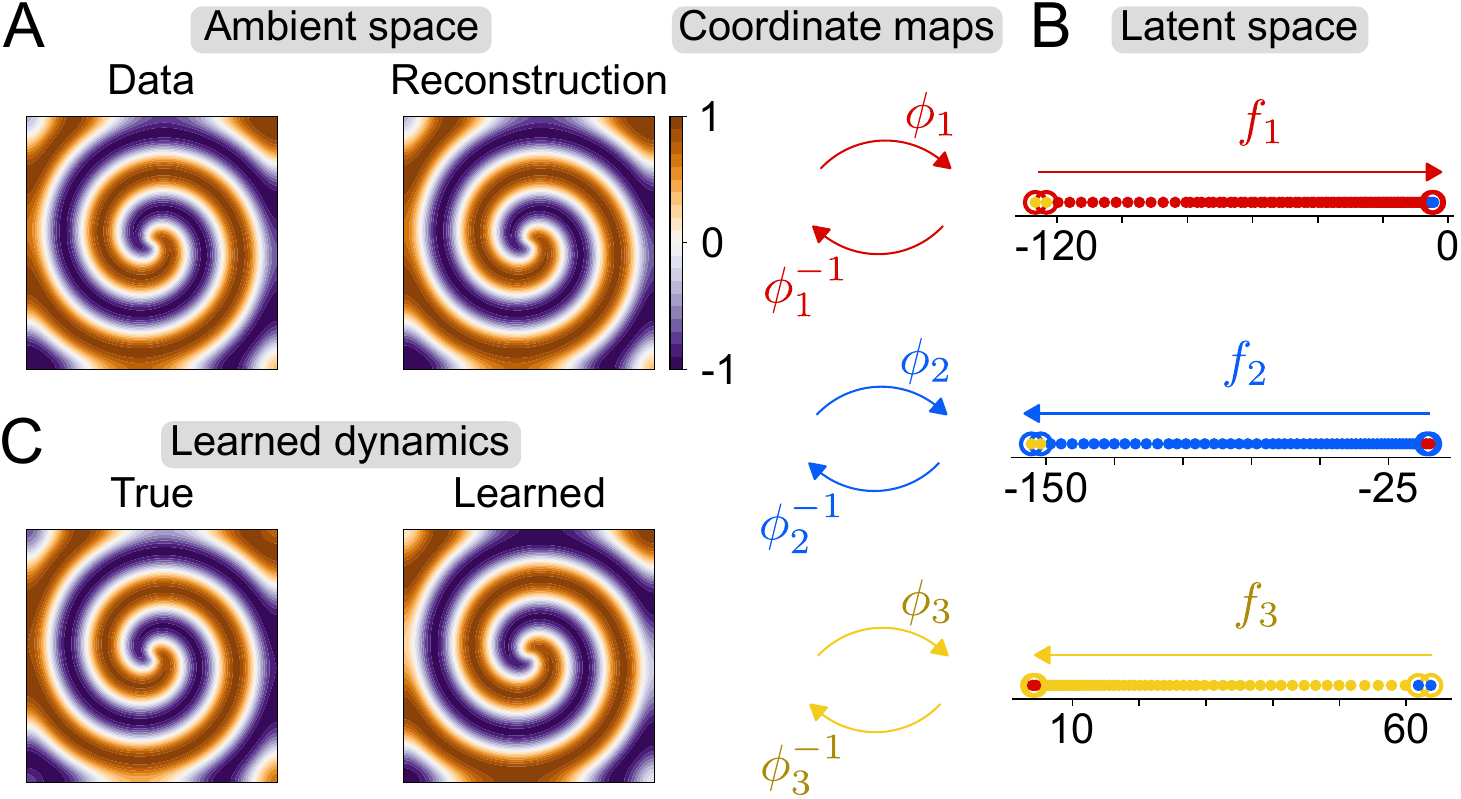}	
\caption{Analogous to figure~\ref{fig:method}, but for a spiral wave from a lambda-omega reaction-diffusion system. In A we show a snapshot of the data and its reconstruction from a one-dimensional representation. In B we show the one-dimensional coordinates in three charts. In C we compare our one-dimensional model's prediction to the true state at 250 time units. }
\label{fig:reactDif}
\end{figure}

We place the data into three clusters and follow our procedure to create an atlas of three charts and a local one-dimensional dynamical model for each chart (figure~\ref{fig:reactDif}A--B). Figure~\ref{fig:reactDif}A shows the reconstruction of one snapshot from its one-dimensional representation. For five models, the average mean squared error was $2.82 \times 10^{-6}$ when using three charts, demonstrating excellent reconstruction. 

We evolve an initial condition forward in time under our learned dynamical model out to 250 time units, far beyond the 10 time units of training. The model's prediction is compared to the ground truth in figure~\ref{fig:reactDif}C. The two are nearly indistinguishable, with the model producing a small error in the phase speed (roughly 0.4\%) that can be improved with further training. This example demonstrates that our method works in higher dimensions, producing a minimal one-dimensional dynamical model for nominally 20402-dimensional data.

\subsection{Kuramoto-Sivashinsky bursting dynamics}	
\label{sec:ksburst}

Our final example comes from the Kuramoto-Sivashinsky (K-S) equation,
\begin{equation}
  \label{eq:ksburst1}
  u_t + uu_x + u_{xx} + \nu u_{xxxx} = 0,
\end{equation}
with $\nu = \frac{16}{71}$ \cite{kirby1992reconstructing}. The K-S equation is used as a model for several physical phenomena, including instabilities in flame fronts, reaction-diffusion systems, and drift waves in plasmas. We discretize the PDE on a grid with 64 points. The dynamics and state space structure are far more complicated than in the previous example. After transients have died out, we obtain complicated bursting dynamics, shown in figure~\ref{fig:ksburst}A. The field bursts between pseudo-steady cellular states of opposite sign. The projection onto the leading spatial Fourier modes clarifies that the state seems to switch between two saddle points that are connected by four heteroclinic orbits \cite{kevrekidis1990back}, forming what may be described as a skeletal attractor. The trajectory is non-periodic, as the state travels along each heteroclinic orbit pseudo-randomly with equal probability. 

\begin{figure}
\centering
\includegraphics[width=0.85\linewidth]{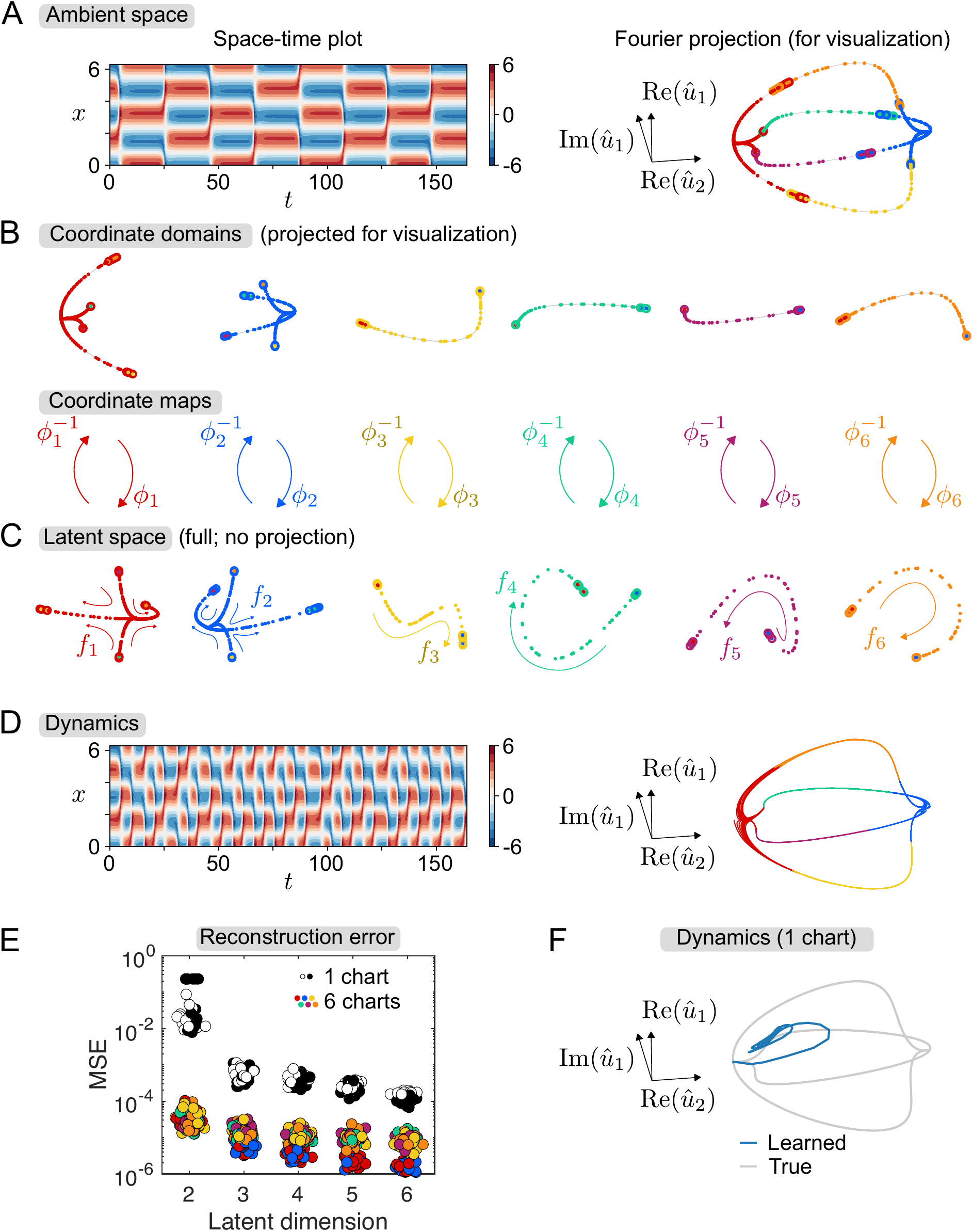}
\caption{\scriptsize{Analogous to figure~\ref{fig:method}, but for bursting data from the K-S system. In A and D, we show space-time plots and projections onto the real part of the second spatial Fourier mode and real and imaginary parts of the first spatial Fourier mode. Note that B shows a projection of the data, whereas C shows the data in the full learned three-dimensional spaces. In A-C, we subsampled the data for visual clarity. (E) Mean squared errors (MSEs) of autoencoders of various bottleneck dimensions when trained on the bursting K-S data, using six charts (coloured markers) and one chart (black and white markers). Autoencoders corresponding to black markers have the same architecture as those used in the six-chart case. Autoencoders corresponding to white markers have approximately the same number of trainable parameters as all six autoencoders in total from the six-chart case. For all cases, 20 trials were performed. (F) Dynamics produced by a one-chart model with a six-dimensional latent space. }}
\label{fig:ksburst}
\end{figure}

Our data cover each heteroclinic orbit four times, with the first half shown in figure~\ref{fig:ksburst}A. The data live on a submanifold of unknown dimension. We will use our method to find the dimension of the submanifold, and compare the results to a one-chart model. 

We place the data into six clusters and follow our procedure to create an atlas of six charts and a local low-dimensional dynamical model for each chart (figure~\ref{fig:ksburst}A--C). We chose to use six charts based on the state space structure in figure~\ref{fig:ksburst}A. The clusters respectively constitute the two saddle points and the four heteroclinic orbits; this clustering of the data was done automatically by $k$-means. 


We successfully built a model with three-dimensional latent spaces. We used a dimension of three since the reconstruction error plateaus there  (see figure~\ref{fig:ksburst}E). It is also evident in figure~\ref{fig:ksburst}E that our multi-chart method captures the submanifold better than one-chart models, attaining significantly lower reconstruction errors. There, we show the reconstruction errors of several autoencoders (using the method in \cite{linot2020deep}) as a function of the dimension of the latent space. The key comparison to make is between the six-chart atlas with three-dimensional latent spaces, and the one-chart atlas with a six-dimensional latent space. Since the submanifold is three-dimensional, the strong Whitney embedding theorem states that it can be smoothly embedded in $\mathbb{R}^6$, i.e., it can be captured by a one-chart atlas with a six-dimensional latent space. Nevertheless, the reconstruction error of the six-chart atlas is an order of magnitude lower than that of the one-chart atlas with twice the dimension and same number of trainable parameters. It is clear that our multi-chart method provides practical benefits even beyond what is theoretically expected. 

In order to recreate qualitatively correct dynamics, we had to use the method described in~\cite{linot2020deep} for the autoencoders, and normalize the local coordinates before learning the dynamics. The normalization we used is similar to whitening, with full details in the Supplementary Information. Additionally, we also had to use a different dataset with off-attractor data to train the neural networks for dynamics; details are in the Supplementary Information. The reason for using off-attractor data to learn the dynamics is that the attractor is very thin (see figure~\ref{fig:ksburst}C), making it difficult to learn accurate dynamics in the three-dimensional latent spaces. When only using the original dataset to learn the dynamics, we found that the learned dynamics often included a stable fixed point that was off the attractor (in a region where there was no data to learn from). The new dynamics dataset was partitioned and transformed to the latent spaces based on the previously trained charts before training the neural networks for the dynamics. 

Even with the above modifications, many of our learned models display periodic dynamics. We have found models that have one stable limit cycle (essentially consisting of two of the heteroclinic orbits), two stable limit cycles (each consisting of an independent pair of the heteroclinic orbits), and one longer stable limit cycle (consisting of all four heteroclinic orbits), in addition to models with non-periodic dynamics. We attribute this to the presence---in the space of weights for the neural networks for the dynamics---of a gluing bifurcation \cite{gambaudo1984collage, glendinning1988new, graham1993pulses}. In short, perturbations to the dynamics can ``glue" the heteroclinic orbits together, forming periodic orbits of various degrees of complexity. It may be possible to avoid the gluing bifurcation by enforcing some of the symmetries present in the K-S system~\cite{kevrekidis1990back}, but we do not pursue this avenue here. Figure~\ref{fig:ksburst} and the discussion below describe results from a model that displays qualitatively correct non-periodic dynamics. 

In figure~\ref{fig:ksburst}D, we show a trajectory that results from evolving an initial condition forward under our learned three-dimensional dynamical model. We are able to obtain qualitatively correct non-periodic bursting dynamics. The quasi-steady cellular states quantitatively match those in the original data. The only quantitative discrepancy is in how long the state stays in a quasi-steady cellular state. This difference can be understood from the fact that these quasi-steady states are approaches to saddle points. The time that a trajectory spends near a saddle point scales as $-\ln y$ near the saddle point, where $y$ is the distance from the stable manifold of the trajectory. The logarithmic divergence explains the sensitivity of the quiescent periods to model error. 

For comparison, in figure~\ref{fig:ksburst}F we show a typical trajectory that results from a one-chart model with a six-dimensional latent space. The state settles onto a fixed point in the model that is off the true attractor, not reflective of the true dynamics. Although higher-dimensional one-chart models are in principle capable of capturing the attractor (by the strong Whitney embedding theorem), as a practical matter, we find that they are unable to capture the qualitative dynamics of this complex system. 

\section{Discussion}
\label{sec:conc}	

We have introduced a method that is able to learn minimal-dimensional dynamical models directly from high-dimensional time series data. Our approach first finds and parameterizes the low-dimensional manifold on which the data live, then learns the dynamics on that manifold. 

Taking inspiration from the rigorous mathematical theory of manifolds, we characterize learning a manifold as learning an atlas of charts: a collection of overlapping patches on the manifold together with invertible maps between points on the patches and points in the corresponding latent spaces. We implement an atlas of charts by clustering the data in state space to form patches, assigning a deep autoencoder to each patch for the dimension reduction step, and using deep neural networks to learn the dynamics in each patch. In our implementation, the charts are mutually independent, making the training process embarrassingly parallelizable. Finally, we sew the local models together to obtain a global dynamical model. We call our method CANDyMan, short for charts and atlases for nonlinear data-driven dynamics on manifolds. 

By examples, we have shown that CANDyMan can learn accurate dynamical models whose dimension is equal to the intrinsic dimensionality of the system. This ability is enabled by the use of multiple charts; single-chart methods are incapable of producing accurate models of intrinsic dimensionality irrespective of the amount of training data, network architecture, etc. In addition, our multi-chart method outperforms an otherwise equal single-chart method even under conditions when the strong Whitney embedding theorem suggests the two should perform equally well. We speculate this is because it is easier to learn the local geometry of a data manifold than the global geometry. 

More broadly, a significant limitation of single-chart methods appears when considering data sets with strongly nonuniform distributions of points. This situation arises naturally in dynamical systems that have features on disparate time scales. For example, in our final example above, the system spends most of its time in regions of state space near the saddle points, with infrequent and brief excursions, often associated with extreme events, to other parts of state space \cite{graham1996alternative}.  If the system is sampled at constant time intervals, then the excursions will form only a small part of the data set, and accordingly have only a small weight in the loss function that is used for training a single-chart manifold representation. Consequently, these excursions, which play a central role in the overall dynamics, may be approximated quite poorly because the large local error from these regions is ``diluted" by the contributions from the more frequently sampled regions. By considering multiple local descriptions of the state space geometry and dynamics, which are trained independently from one another, the method we describe here avoids this dilution problem.

Our approach stands in stark contrast to recently developed methods based on Koopman operator theory \cite{takeishi2017learning, lusch2018deep, otto2019linearly} and reservoir computing \cite{pathak2017using, pathak2018model, vlachas2020backpropagation}. Those methods can produce highly accurate data-driven dynamical models, but they require (explicitly or implicitly) high-dimensional state representations---larger than the dimension of the full state space---in order to be accurate. The purpose of CANDyMan, on the other hand, is to enable state representations of minimal dimensionality. 


A question that naturally arises is how many charts must be used. This question is intimately tied to the dimension that we are able to reduce the data to. We may always use a single chart, but as we have shown, doing so limits the dimension reduction that can be achieved. Moreover, we also showed that one-chart representations can be poor. How many charts do we need in order to obtain a minimal-dimensional representation of the data, and how do we know what the minimal dimension is? For now, we suggest an empirical approach where one would sample in the space of number of charts and number of dimensions. It may be easier to first find the appropriate dimension by only considering a low number of charts that cover small incomplete patches of the data, and tuning the dimension of these charts; since the dimension of a manifold is a global property, it can be established locally in this way. Several techniques can provide an initial estimate of the dimension \cite{camastra2016intrinsic}. Once the appropriate dimension is found, the number of charts can be tuned empirically. Tools of graph theory and topological data analysis may provide guidance into the choice of the number of charts---this will be an interesting topic of future work.

An issue that deserves consideration in future work is the structure that should be imposed on the coordinate maps. Although the theory of manifolds is agnostic to the form that coordinate maps take (as long as they are homeomorphic), we saw in our last example that imposing structure can be helpful from a practical standpoint. One possible structure is that, in addition to being homeomorphic, the coordinate maps should also be isometric \cite{tenenbaum2000global}. Another possible structure would be one that induces some level of interpretability, as in \cite{champion2019data}. The appropriate structure would be one that makes learning the correct dynamics easier. 


Besides enabling the analysis and prediction of a system's behaviour, CANDyMan also has the potential to reveal hidden properties of the system. 
For example, the presence of conserved quantities in a system will reduce the dimension of the solution manifold beyond what a naive representation of the system would yield.
In cases where the hidden properties of a system are not obvious, the intrinsic variables provided by CANDyMan would need to be translated to interpretable physics, and we suggest this as a future research direction. 

Finally, because our method decomposes state space into patches, we believe that it is particularly suited to applications that display very different behaviours in different parts of state space, such as the bursting and intermittency characteristic of turbulent fluid flows, neuronal activity, and solar flares, among others.

\textit{Acknowledgements:} We acknowledge the use of the Sabine cluster from the Research Computing Data Core at the University of Houston, and the assistance of Dr. Deepak A. Kaji and the Luke cluster. 

\textit{Data access:} Data and code have been deposited in GitHub, available at \\ \href{https://github.com/dfloryan/CANDyMan}{https://github.com/dfloryan/CANDyMan}.

\textit{Author contributions:} DF and MDG designed research, performed research, analyzed data, and wrote the paper. 

\textit{Competing interests:} The authors declare no competing interests.

\textit{Funding:} This work was supported by Air Force Office of Scientific Research grant FA9550-18-0174 and Office of Naval Research grant N00014-18-1-2865 (Vannevar Bush Faculty Fellowship).

\bibliography{references} 

\pagebreak
\widetext
\begin{center}
\textbf{\large Supplementary Information for Floryan and Graham, Data-driven discovery of intrinsic dynamics}
\end{center}
\setcounter{section}{0}
\setcounter{equation}{0}
\setcounter{figure}{0}
\setcounter{table}{0}
\setcounter{page}{1}
\setcounter{footnote}{0}
\makeatletter
\renewcommand{\thesection}{S\arabic{section}}
\renewcommand{\theequation}{S\arabic{equation}}
\renewcommand{\thefigure}{S\arabic{figure}}
\renewcommand{\bibnumfmt}[1]{[S#1]}

\section{Limitations of global manifold learning methods}
\label{sec:manFail}

As discussed in the main text, global manifold learning methods can produce discontinuities or coordinate maps that are not one-to-one. The main text demonstrates this effect for an autoencoder, and here we additionally demonstrate the same effect for a suite of manifold learning methods: kernel principal component analysis (PCA) \cite{scholkopf1998nonlinear}, Isomap \cite{tenenbaum2000global}, Laplacian eigenmaps \cite{belkin2003laplacian}, Hessian eigenmaps \cite{donoho2003hessian}, locally linear embedding \cite{roweis2000nonlinear}, and t-SNE \cite{van2008visualizing}. We use the methods as implemented in the Scikit-learn library \cite{pedregosa2011scikit}. 

In figure~\ref{fig:circFailPopular}, we show the one-dimensional latent representations of 20 evenly spaced points on the unit circle produced by applying the above manifold learning methods to the two-dimensional Cartesian coordinates of the points. In every case, points of dissimilar colours are mixed together. As discussed in the main text, this happens because the coordinate maps are not one-to-one, mapping different parts of the circle to the same point in the latent spaces. Intuitively, these manifold learning methods reduce the dimension of the data from two to one by essentially collapsing the circle to a dimension of one. Since the circle is a closed object, it collapses onto itself, leading to the observed effect. 

Note that the effect shown in figure~\ref{fig:circFailPopular} is purely due to the geometry of the circle, independent of the values of the hyperparameters of each method (which we verified). We have chosen values for the hyperparameters that produce visually clear results. For kernel PCA, we used a radial basis function kernel with $\gamma = 10$. For Isomap, we used two neighbours. For locally linear embedding, we used two neighbours. For Laplacian eigenmaps, we used four neighbours. For Hessian eigenmaps, we used five neighbours. For t-SNE, we set perplexity equal to five. All other parameters were set to their default values. For meanings of these parameters, we refer the reader to the documentation of the Scikit-learn library. 

\begin{figure}
\centering
\includegraphics[width=0.7\linewidth]{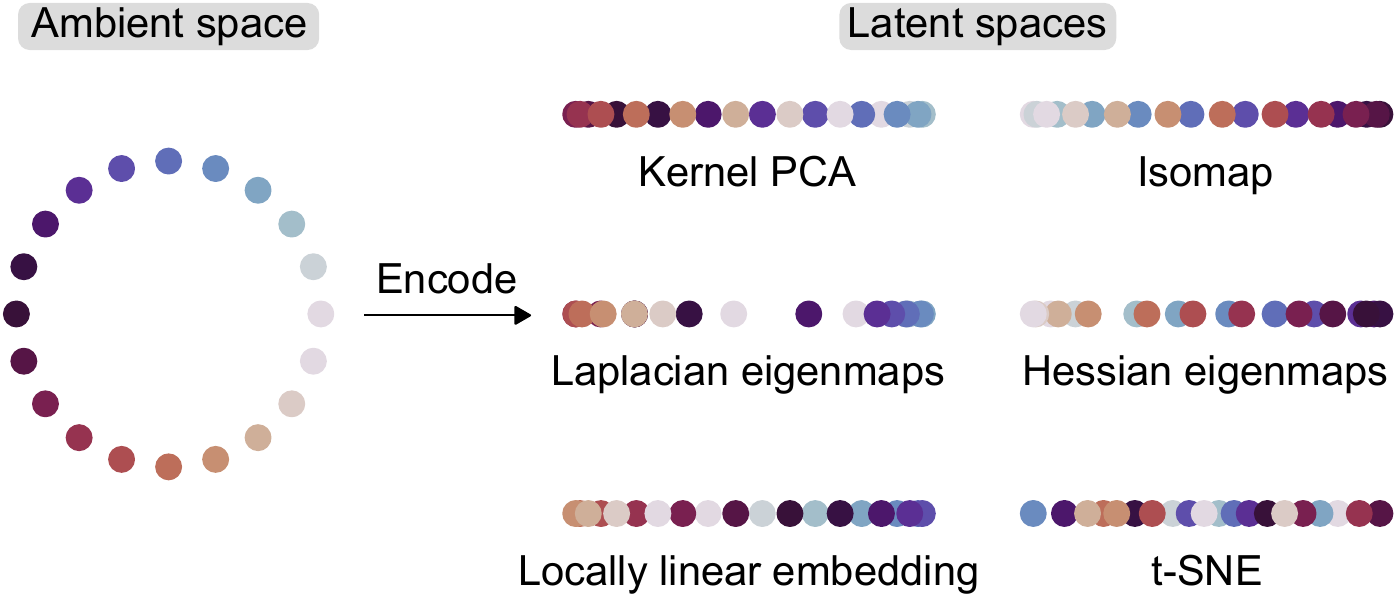}
\caption{One-dimensional latent representations of points on a circle produced by several manifold learning methods. }
\label{fig:circFailPopular}
\end{figure}

\section{Comparing CANDyMan to single-chart methods}
\label{sec:comp}

As demonstrated in the main text and previous section, global---that is, single-chart---methods are unable to produce accurate models of intrinsic dimensionality. A contemporaneous work has proposed a method to learn dynamical models of intrinsic dimensionality~\cite{chen2021discovering}; it is a one-chart method. That work and ours both use the reaction-diffusion system as an example, so we compare the results of the two methods here. We thank the authors of~\cite{chen2021discovering} for making their code and data available online\footnote{https://github.com/BoyuanChen/neural-state-variables}. 

Using their code and data\footnote{We used the second half of their data, by which time the system settles on a limit cycle.}, we learned a two-dimensional model for the reaction-diffusion system (as done in the original work), as well as a one-dimensional model, only changing the dimension of the latent space. Note that the data consist of $128 \times 128$ image frames from a movie. We found that the two-dimensional model produced trajectories nearly indistinguishable from the ground truth test data, but the one-dimensional model was inaccurate, producing sharp jumps. We then used our multi-chart method---using the same architecture, training procedure, and training data as in \cite{chen2021discovering}, with three charts---to learn a one-dimensional model, and found it to produce trajectories nearly indistinguishable from the ground truth. The mean squared errors of the trajectories produced by the three models are shown in figure~\ref{fig:rdifcomp}A. Supplementary Movie M1 compares all three models to the ground truth.

Pixel-wise errors better represent the difficulties that a one-chart one-dimensional model encounters. In figure~\ref{fig:rdifcomp}B, we show the intensity of the green channel (normalized to be between zero and one) of one of the pixels during the first 100 time steps of a trajectory. The single-chart one-dimensional model is inaccurate, producing wild jumps, whereas our multi-chart one-dimensional model is accurate. Four consecutive frames produced by the ground truth dynamics and the single-chart one-dimensional model, occurring near the first sharp jump in figure~\ref{fig:rdifcomp}B, are shown in figures~\ref{fig:rdifcomp}C and D, respectively. For ease of comparison, we also show skeletal images below each frame, produced by only colouring pixels of a certain intensity. Pixels have been coloured red in regions that differ significantly from the ground truth results. The single-chart two-dimensional model and multi-chart one-dimensional model produce results that are nearly indistinguishable from the ground truth. 

\begin{figure}
\centering
\includegraphics[width=0.6\linewidth]{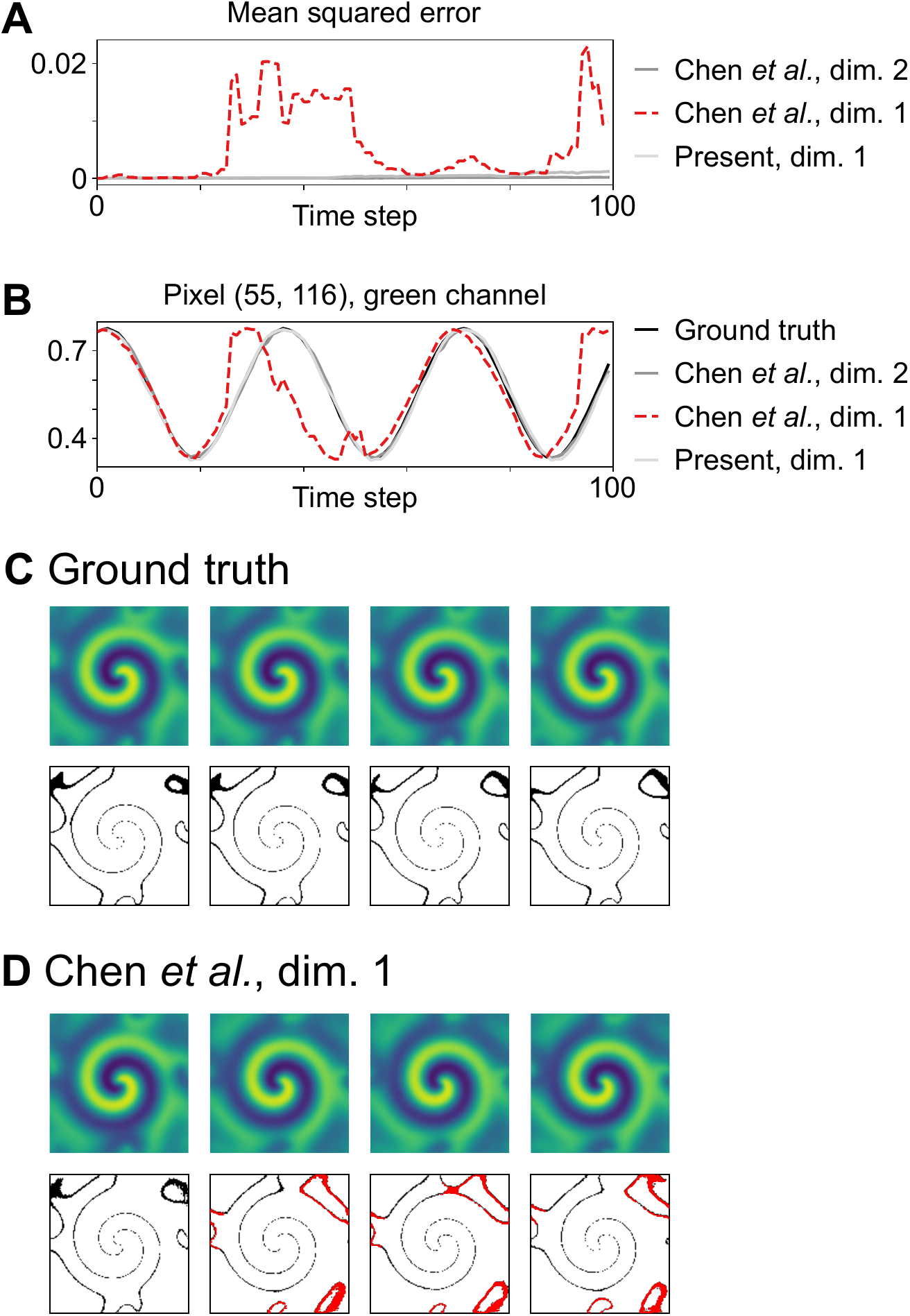}
\caption{Comparison between ground truth and various models for a reaction-diffusion system. (A)~Mean squared error as a function of time. (B) Intensity of the green channel of pixel (55, 116). Only the multi-chart method proposed in the present work produces an accurate one-dimensional model. (C) Four consecutive snapshots from the ground-truth trajectory around 25 time steps from the initial condition. (D) Corresponding snapshots from a one-chart one-dimensional model using the method described in \cite{chen2021discovering}. The skeletal pictures highlight features of the trajectory and are coloured red when differing significantly from the ground truth. }
\label{fig:rdifcomp}
\end{figure}

The work of \cite{champion2019data} also uses the same example, producing a two-dimensional model. Their method is also a single-chart method. Clearly, multiple charts are generally needed to produce accurate dynamical models of intrinsic dimensionality. We believe our CANDyMan framework can be fruitfully combined with the works of \cite{champion2019data} and \cite{chen2021discovering} to produce even more accurate and interpretable models that are truly of intrinsic dimensionality.

\section{Additional examples}
\label{sec:addEx}

Here, we demonstrate the CANDyMan method on additional examples not included in the main text.

\subsection{Periodic dynamics on a torus}	
\label{sec:torper}

Consider a particle moving along the surface of a torus. The particle travels at constant speeds in the poloidal and toroidal directions, three times as fast in the poloidal direction as in the toroidal direction, generating a periodic orbit. The data live on a one-dimensional submanifold of $\mathbb{R}^3$, shown in figure~\ref{fig:torus1d}A. 

\begin{figure}
\centering
\includegraphics[width=\linewidth]{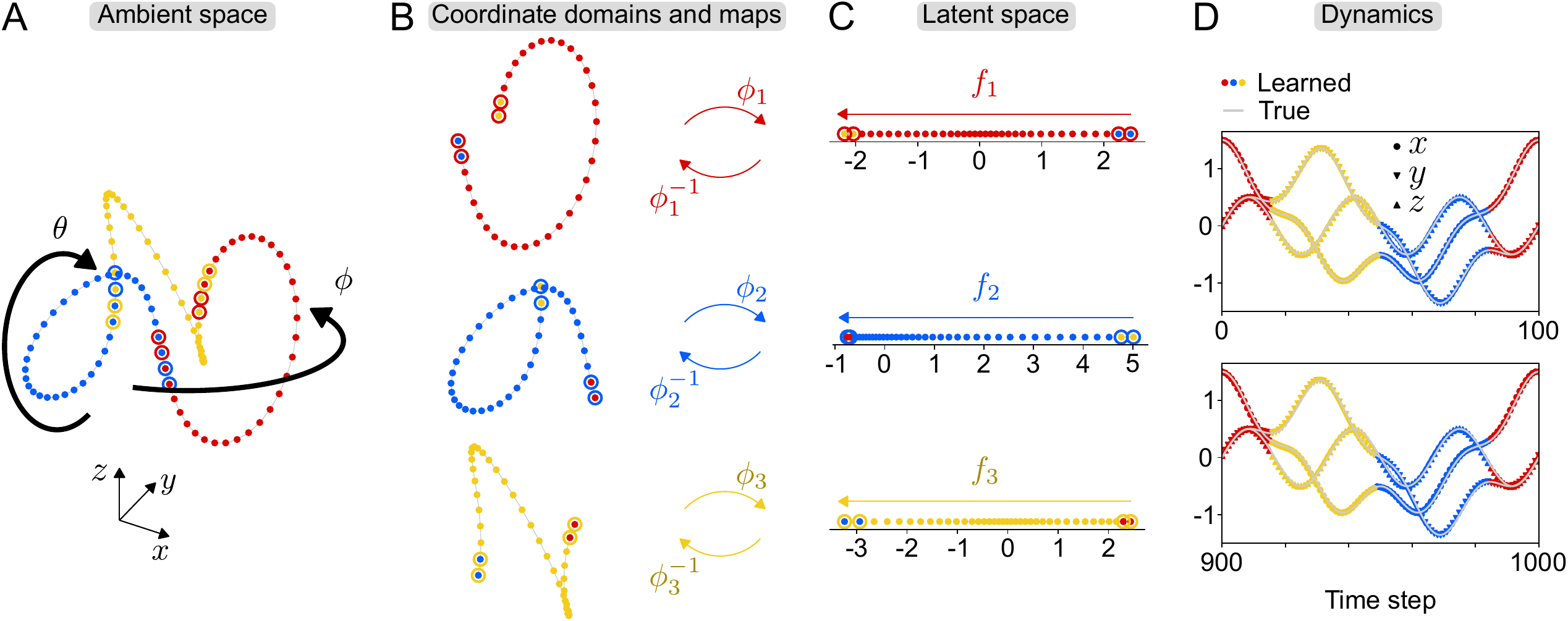}
\caption{(A) Data from a periodic orbit on the surface of a torus. (B) The learned coordinate domains. (C) The learned latent spaces/local coordinates. (D) Comparison between the learned dynamics and true dynamics. The first (top) and tenth (bottom) cycles of motion are well predicted.}
\label{fig:torus1d}
\end{figure}

We place the data into three clusters and follow our procedure to create an atlas of three charts and a local low-dimensional dynamical model for each chart (figure~\ref{fig:torus1d}A--C). The resulting global dynamical model is excellent. In figure~\ref{fig:torus1d}D, we show that the learned dynamics (coloured markers) match the true dynamics (grey curves) closely. Note especially the seamless transitions between charts.

\subsection{Kuramoto-Sivashinsky beating dynamics}
\label{sec:ksbeat}

Our next example comes from the Kuramoto-Sivashinsky (K-S) partial differential equation, 
\begin{equation}
  \label{eq:ks}
  u_t + uu_x + u_{xx} + \nu u_{xxxx} = 0,
\end{equation}
for $0 \le x \le 2\pi$, with periodic boundary conditions. We set $\nu = \frac{16}{337}$ \cite{rowley2000reconstruction} and compute a numerical solution on a grid of 64 points. After transients have died out, we obtain a beating standing wave. The data live on a one-dimensional submanifold of $\mathbb{R}^{64}$, shown in figure~\ref{fig:KSbeating}A.

\begin{figure}
\centering
\includegraphics[width=0.6\linewidth]{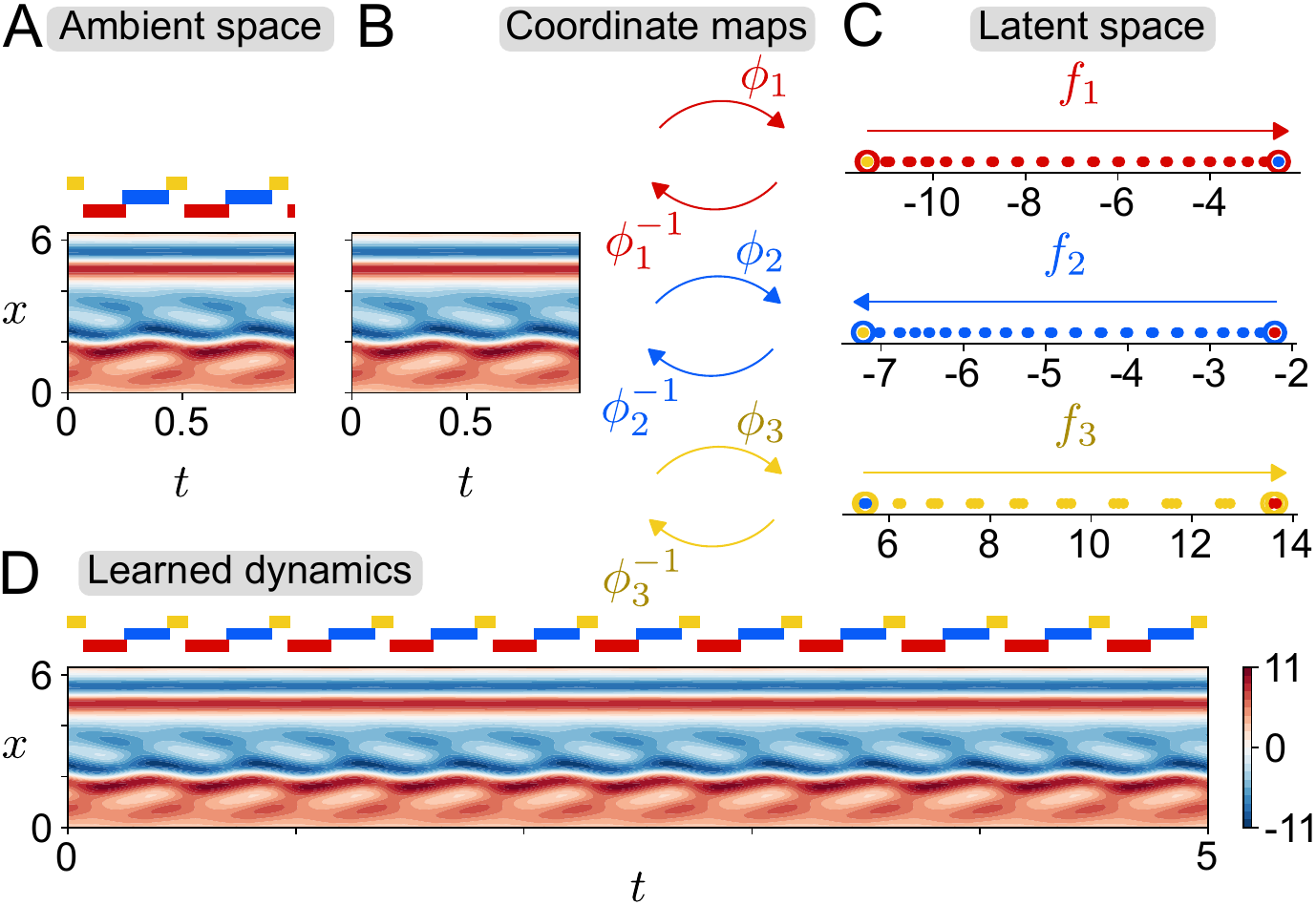}	
\caption{(A) Space-time plot of data. The coloured bars show which learned coordinate domain the state is in at each instant in time. (B) Reconstruction of data from the learned latent spaces/local coordinates. (C) The learned latent spaces/local coordinates. (D) The learned dynamics. }
\label{fig:KSbeating}
\end{figure}

We place the data into three clusters and follow our procedure to create an atlas of three charts and a local low-dimensional dynamical model for each chart (figure~\ref{fig:KSbeating}A--C). Figure~\ref{fig:KSbeating}B shows the reconstruction of the data from their local coordinates, demonstrating excellent reconstruction. 

In figure~\ref{fig:KSbeating}D, we show a trajectory that results from evolving an initial condition forward in time under our learned dynamical model. The transitions between charts are seamless and the trajectory is indistinguishable from the data. This example demonstrates that our method works in higher dimensions, producing a minimal one-dimensional dynamical model for nominally 64-dimensional data.

\subsection{Kuramoto-Sivashinsky beating travelling dynamics}
\label{sec:ksbt}

Our next example also comes from the K-S equation, this time with $\nu = \frac{4}{87}$ \cite{rowley2000reconstruction}. After transients have died out, we obtain a beating travelling wave. The travelling period and beating period are incommensurable, making the orbit quasiperiodic. Note the separation in timescales: the two periods differ by a factor of nearly 200. This quasiperiodic orbit lives on a two-dimensional submanifold (a 2-torus) of $\mathbb{R}^{64}$, shown in figure~\ref{fig:ksbtphase}A. 

\begin{figure}
\centering
\includegraphics[width=0.8\linewidth]{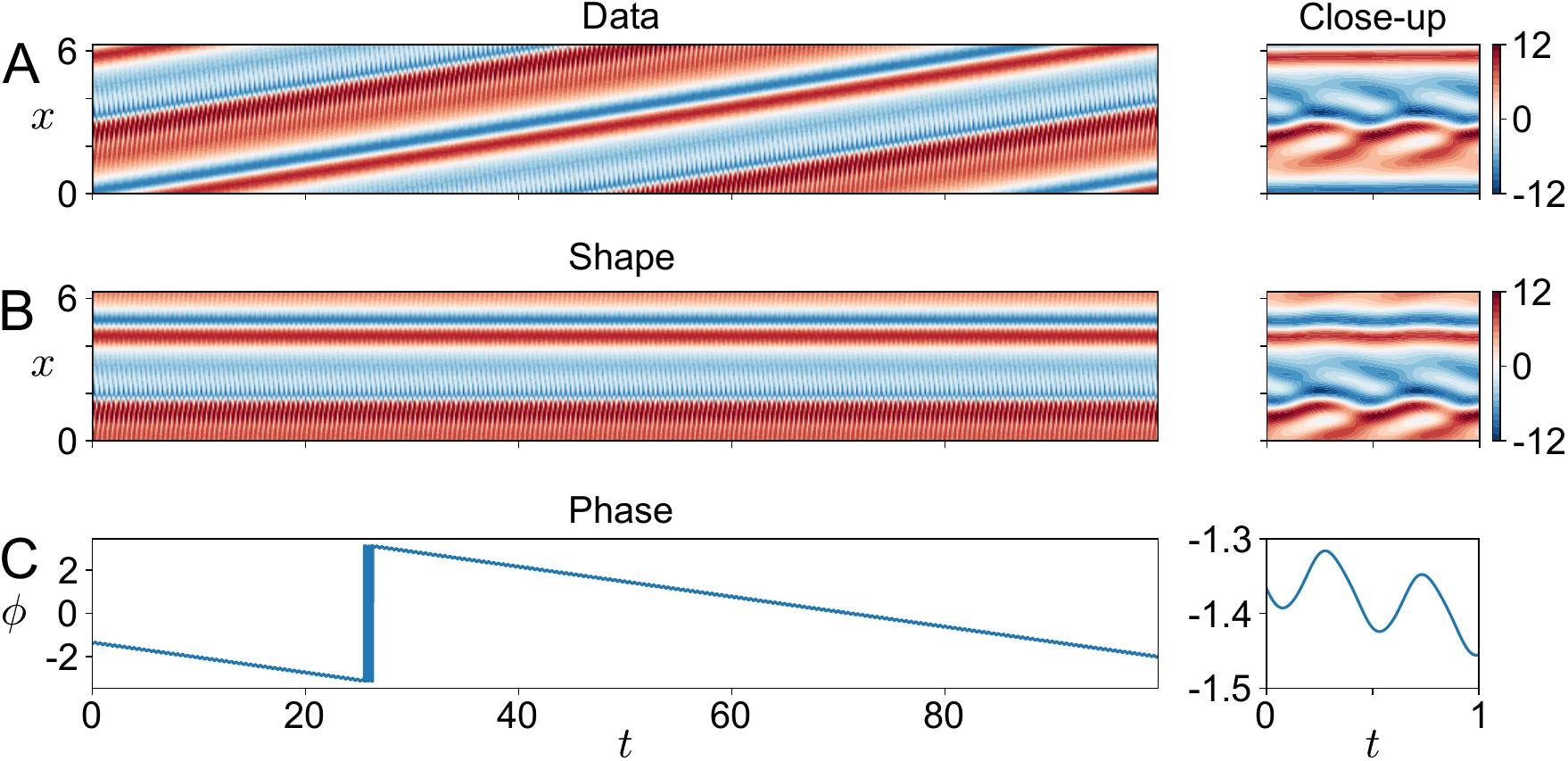}	
\caption{The beating travelling wave data from the K-S system (A) are separated into shape (B) and phase (C) data. Plots on the right show close-up views of the first time unit of the dynamics.}
\label{fig:ksbtphase}
\end{figure}

We separate the data into shape (figure~\ref{fig:ksbtphase}B) and phase (figure~\ref{fig:ksbtphase}C) variables (the shape variable's first Fourier mode's phase is zero) \cite{budanur2015reduction, linot2020deep}. Because of the translational equivariance of~\eqref{eq:ks}, the dynamics only depend on the shape variable, which lives on a one-dimensional submanifold of $\mathbb{R}^{64}$. 

Our training data are shown in figure~\ref{fig:ksbt}A. The data span a bit more than two beating periods, just over 1\% of the travelling period. We place the data into three clusters and follow our procedure to create an atlas of three charts and a local low-dimensional dynamical model for each chart (figure~\ref{fig:ksbt}A--C). Additionally, for each chart we train a neural network that takes the shape variable (in local coordinates) as input and predicts the change in phase over one time step, giving the phase dynamics \cite{linot2020deep}. 

\begin{figure}
\centering
\includegraphics[width=0.75\linewidth]{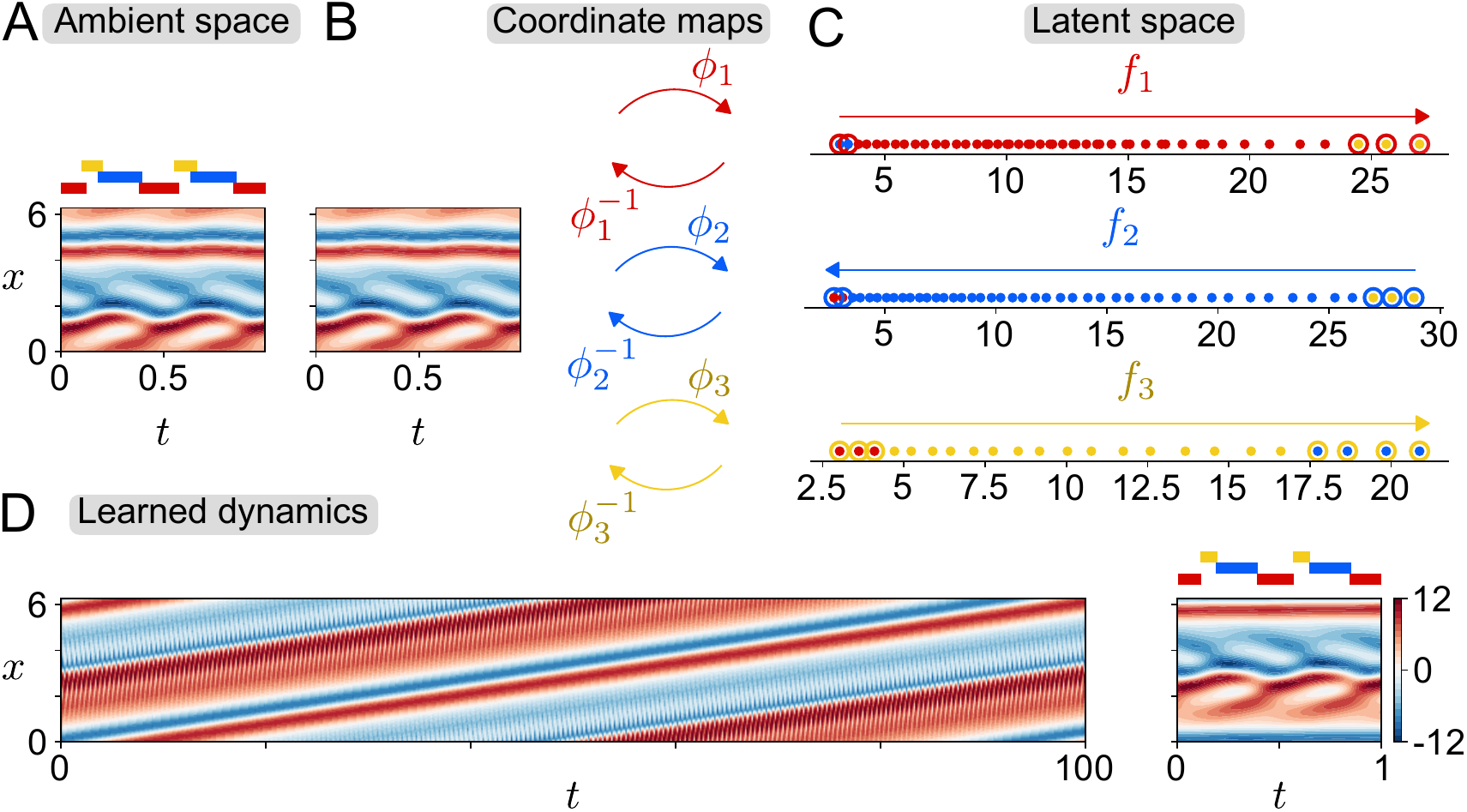}	
\caption{Analogous to figure~\ref{fig:KSbeating}, but for a beating travelling wave. In D, we show the predicted dynamics starting from the same initial condition as in figure~\ref{fig:ksbtphase}A. }
\label{fig:ksbt}	
\end{figure}

In figure~\ref{fig:ksbt}D, we show a trajectory that results from evolving an initial condition forward in time under our learned dynamical model. Comparing figure~\ref{fig:ksbt}D to figure~\ref{fig:ksbtphase}A, our learned model again produces good results with seamless transitions between charts. The beating period of our model is $0.456 \pm 0.001$ time units, identical to the data, and the travelling period of our model is $94.88 \pm 0.10$ time units, compared to $90.46 \pm 0.10$ time units for the data. The discrepancy improves with increased training time of the neural networks.

It is worth re-iterating that the two periods of the quasiperiodic data differ by a factor of almost 200. Despite the large separation in timescales, our method produced a good dynamical model. Moreover, we were able to take advantage of a symmetry present in the system to learn from data that spans a small fraction of one of the timescales while nevertheless capturing that timescale in the final dynamical model.

\section{Details of datasets and models}
\label{sec:det}

Here we provide details about the training datasets, neural network architectures, training procedures, and hyperparameters used for each of the examples. Neural networks were built and trained using Keras \cite{chollet2015keras}. Unless otherwise indicated, all neural network layers are fully-connected, the mean squared error is used as the loss function, the default Adam optimizer is used for training \cite{kingma2014adam}, the default Glorot uniform initializer is used for weight initialization \cite{glorot2010understanding}, and the batch size is the full number of training data points in the corresponding chart. 

We did not attempt to optimize architectures nor hyperparameters, simply using ones that produced accurate results. To choose the dimension of the latent space, we either knew what the true dimension of the submanifold was (in which case we used that dimension), or we based our decision on reconstruction errors (i.e., figure~6E in the main text suggests that the latent dimension of the bursting data is 3 since the reconstruction error reaches a plateau there \cite{linot2020deep}; note that the latent dimension cannot be 2 since such a low dimension is not capable of producing the dynamics we observe). To choose the number of charts, we often knew the minimum number needed, and chose nearly that many. For the bursting example, we chose to use 6 charts based on an exploratory data analysis, where we saw that the state space structure of the data consists of two saddles connected by four heteroclinic orbits.

\subsection{Examples from main text}
\label{sec:mtdet}

\subsubsection{Periodic dynamics on a circle}
\label{sec:circSI}

The training data are the Cartesian $(x, y)$ coordinates of a particle as it moves counterclockwise around the unit circle at a constant speed. There are 40 data points in total, uniformly spaced $\pi/20$ radians apart on the unit circle. 

We use three charts (three clusters in $k$-means). The graph is built by connecting points to their two nearest neighbours. Clusters are expanded twice along the graph. 

The neural network parameters are listed in table~\ref{tab:circSI}. The autoencoders are trained with a learning rate of 0.01, and the neural networks for dynamics are trained with a learning rate of 0.005.

\begin{table}
\centering
\begin{tabular}{c | c | c | c} 
  \hline
  & Shape & Activations & Epochs \\
  \hline
  Encoders & 2 : 32 : 32 : 16 : 4 : 1 & elu : elu : elu : elu : linear & 1000 \\
  Decoders & 1 : 4 : 16 : 32 : 32 : 2 & elu : elu : elu : elu : linear & 1000 \\
  Dynamics & 1 : 32 : 32 : 16 : 4 : 1 & elu : elu : elu : elu : linear & 500 \\
  \hline
\end{tabular}
\caption{Neural network parameters for periodic dynamics on a circle. ``Shape'' indicates the dimension of each layer, ``activation'' indicates the activation functions between layers, and ``epochs'' indicates the number of epochs used for training. }	
\label{tab:circSI}
\end{table}

\subsubsection{Quasiperiodic dynamics on a torus}
\label{sec:torqpSI}

The training data are the Cartesian $(x, y, z)$ coordinates of a particle as it moves around the surface of a torus. The surface of the torus is given by
\begin{equation}	
  \label{eq:torqpSI1}	
  x = (1 + 0.5\cos \theta)\cos \phi,\;
  y = (1 + 0.5\cos \theta)\sin \phi,\;
  z = 0.5\sin \theta,
\end{equation}
where $\theta$ is the poloidal coordinate and $\phi$ is the toroidal coordinate. The particle moves at a constant rate in the poloidal and toroidal directions, $\sqrt{3}$ times as fast in the poloidal direction, generating a quasiperiodic orbit. There are 1000 data points in total, uniformly spaced $3\pi/100$ radians apart in the toroidal direction, covering 15 full periods of motion in the toroidal direction.

We use six charts (six clusters in $k$-means). The graph is built by connecting points to their five nearest neighbours. Clusters are expanded twice along the graph. 

The neural network parameters are listed in table~\ref{tab:torqpSI}. The autoencoders are trained with an exponential decay learning rate schedule (initial learning rate 0.01, decay rate 0.8 every 200 steps using a staircase function), and the neural networks for dynamics are also trained with an exponential decay learning rate schedule (initial learning rate 0.01, decay rate 0.9 every 200 steps using a staircase function). 

\begin{table}
\centering
\begin{tabular}{c | c | c | c} 
  \hline
  & Shape & Activations & Epochs \\
  \hline
  Encoders & 3 : 32 : 64 : 32 : 16 : 4 : 2 & elu : elu : elu : elu : elu : linear & 2000 \\
  Decoders & 2 : 4 : 16 : 32 : 64 : 32 : 3 & elu : elu : elu : elu : elu : linear & 2000 \\
  Dynamics & 2 : 32 : 32 : 16 : 4 : 2 & elu : elu : elu : elu : linear & 2000 \\
  \hline
\end{tabular}
\caption{Neural network parameters for quasiperiodic dynamics on a torus. ``Shape'' indicates the dimension of each layer, ``activation'' indicates the activation functions between layers, and ``epochs'' indicates the number of epochs used for training. }
\label{tab:torqpSI}
\end{table}

\subsubsection{Reaction-diffusion system}
\label{sec:reactDifSI}

We compute a numerical solution to a lambda-omega reaction-diffusion system governed by
\begin{equation}
\begin{aligned}
  \label{eq:reactDifSI1}
  u_t &= [1 - (u^2 + v^2)]u + \beta (u^2 + v^2)v + d_1 (u_{xx} + u_{yy}), \\
  v_t &= -\beta(u^2 + v^2)u + [1 - (u^2 + v^2)] v + d_2 (v_{xx} + v_{yy}),
\end{aligned}
\end{equation}
for $-10 \le x \le 10$, $-10 \le y \le 10$. The parameter values are $d_1 = d_2 = 0.1$ and $\beta = 1$, and we impose homogeneous Neumann boundary conditions on $u$ and $v$. The domain is discretized into a uniform $101 \times 101$ grid, and the second-order derivatives from the diffusion terms are approximated by second-order centered finite differences. Time integration is performed using MATLAB's \texttt{ode45} function with default settings \cite{matlab2021}. The initial condition is 
\begin{equation}
\begin{aligned}
  \label{eq:reactDifSI2}
  u(x, y, 0) &= \tanh \left[ \sqrt{x^2 + y^2} \cos \left( \angle(x + iy) - \sqrt{x^2 + y^2} \right) \right], \\
  v(x, y, 0) &= \tanh \left[ \sqrt{x^2 + y^2} \sin \left( \angle(x + iy) - \sqrt{x^2 + y^2} \right) \right].
\end{aligned}
\end{equation}

Once transients decay, we obtain a spiral wave and collect our data. The training data are the values of the fields $u$ and $v$ at all grid points; our dataset consists of 200 such data points, uniformly spaced 0.05 time units apart, covering a bit more than one period of the spiral wave. One period of the spiral wave is shown in figure~\ref{fig:rdcyc}. 

\begin{figure}
\centering
\includegraphics[width=0.65\linewidth]{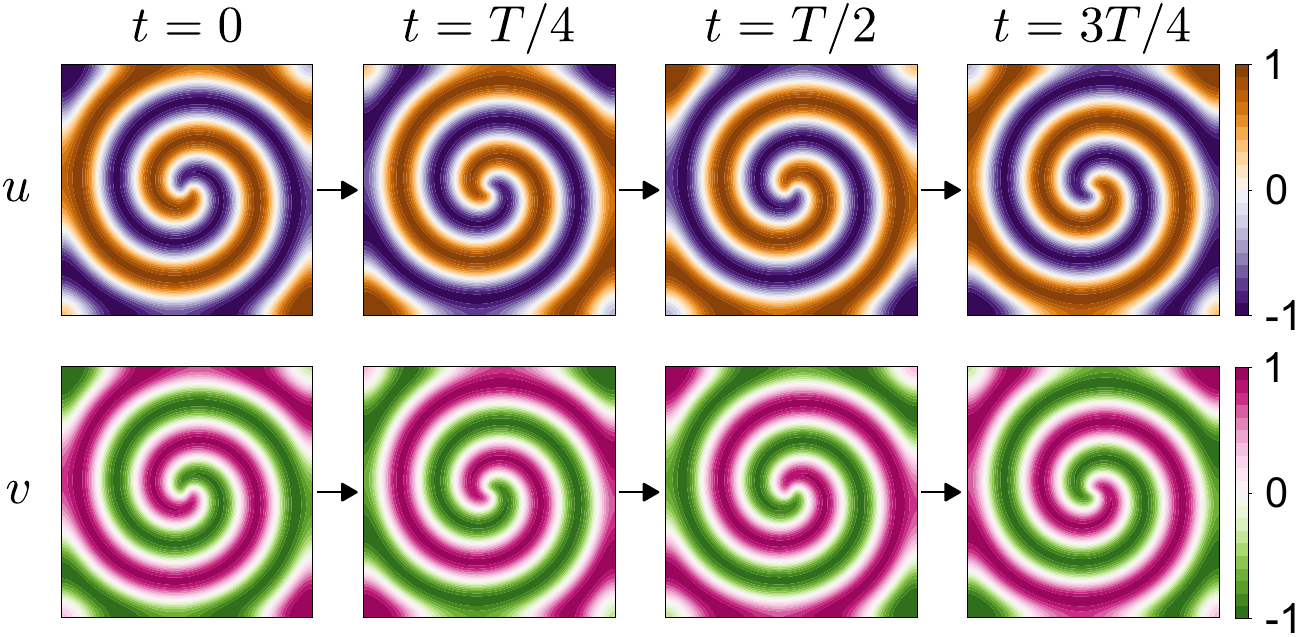}
\caption{Four snapshots showing one period $T$ of the spiral wave. }
\label{fig:rdcyc}
\end{figure}

We use three charts (three clusters in $k$-means). The graph is built by connecting points to their four nearest neighbours. Clusters are expanded once along the graph. 

The neural network parameters are listed in table~\ref{tab:reactDifSI}. The autoencoders are trained with an exponential decay learning rate schedule (initial learning rate 0.01, decay rate 0.8 every 200 steps using a staircase function), and the neural networks for dynamics are also trained with an exponential decay learning rate schedule (initial learning rate 0.01, decay rate 0.9 every 200 steps using a staircase function). 

\begin{table}
\centering
\begin{tabular}{c | c | c | c} 
  \hline
  & Shape & Activations & Epochs \\
  \hline
  Encoders & 20402 : 128 : 64 : 16 : 8 : 1 & elu : elu : elu : elu : linear & 2000 \\
  Decoders & 1 : 8 : 16 : 64: 128 : 20402 & elu : elu : elu : elu : linear & 2000 \\
  Dynamics & 1 : 32 : 32 : 16 : 4 : 1 & elu : elu : elu : elu : linear & 2000 \\
  \hline
\end{tabular}
\caption{Neural network parameters for spiral wave dynamics of the reaction-diffusion system. ``Shape'' indicates the dimension of each layer, ``activation'' indicates the activation functions between layers, and ``epochs'' indicates the number of epochs used for training. }
\label{tab:reactDifSI}
\end{table}

\subsubsection{Kuramoto-Sivashinsky bursting dynamics}
\label{sec:ksburstSI}

We compute a numerical solution to the Kuramoto-Sivashinsky equation using the scheme described in Section~\ref{sec:ksbeatSI}. We set $\nu = \frac{16}{71}$; once transients decay, we obtain bursting dynamics and collect our data. The training data are the values of the field $u$ at 64 points spaced equally in $x$; our dataset consists of 6565 such data points, uniformly spaced 0.05 time units apart, covering 16 bursts. 

The data described above are used to train the autoencoders. The neural networks for dynamics are trained using a different dataset. To create the dynamics dataset, we took every third data point of the autoencoder dataset (2189 in total), perturbed them, and used the resulting data as initial conditions for new simulations. The perturbations are given by
\begin{equation}	
  \label{eq:ksburstSI1}	
  a_1 \cos 2x + a_2 \cos x + a_3 \sin x,	
\end{equation}
where the $a_i$ are randomly sampled from a uniform distribution spanning $[-0.05, 0.05]$. The new simulations were all run for 1.5 time units, and only the last time unit of data was stored. In the end, our dynamics dataset consists of 2189 slightly-off-attractor trajectories, each spanning one time unit and consisting of 21 data points uniformly spaced 0.05 time units apart. 

We use six charts (six clusters in $k$-means). The graph is built by connecting points to their four nearest neighbours. Clusters are expanded twice along the graph. 

The neural network parameters are listed in table~\ref{tab:ksburstSI}. For the autoencoders, we use the method described in \cite{linot2020deep} with $\alpha = 1$ (cf. eq.~4 in that work). The autoencoders are trained with an exponential decay learning rate schedule (initial learning rate 0.01, decay rate 0.8 every 200 steps using a staircase function), and the neural networks for dynamics are also trained with an exponential decay learning rate schedule (initial learning rate 0.01, decay rate 0.9 every 200 steps using a staircase function).

\begin{table}
\centering
\begin{tabular}{c | c | c | c} 
  \hline
  & Shape & Activations & Epochs \\
  \hline
  Encoders & 64 : 128 : 64 : 16 : 8 : 3 & elu : elu : elu : elu : linear & 1000 \\
  Decoders & 3 : 8 : 16 : 64 : 128 : 64 & elu : elu : elu : elu : linear & 1000 \\
  Shape dynamics & 3 : 16 : 64 : 64 : 16 : 3 & elu : elu : elu : elu : linear & 1000 \\
  \hline
\end{tabular}
\caption{Neural network parameters for bursting dynamics of the Kuramoto-Sivashinsky equation. ``Shape'' indicates the dimension of each layer, ``activation'' indicates the activation functions between layers, and ``epochs'' indicates the number of epochs used for training. }
\label{tab:ksburstSI}
\end{table}

In this example, we add an additional step before learning the dynamics, which we referred to as normalization in the main text. In each chart, we take the training data in their local coordinates, subtract the mean, and perform principal components analysis (PCA), changing the local coordinates to their coordinates in the PCA basis (without any dimension reduction). We then normalize each PCA coordinate by the data's standard deviation along the coordinate. This set of steps is similar to whitening. We found this type of normalization to be helpful in learning accurate dynamics. The rationale for normalizing the local coordinates before learning the dynamics is that the saddle points have sharp cusps leading into them, and we thought normalizing the local coordinates would make the cusps less sharp and therefore make the sensitive dynamics easier to learn.

\subsection{Additional examples from Supplementary Information}
\label{sec:sidet}

\subsubsection{Periodic dynamics on a torus}
\label{sec:torperSI}

The training data are the Cartesian $(x, y, z)$ coordinates of a particle as it moves around the surface of a torus. The surface of the torus is given by~\eqref{eq:torqpSI1}. The particle moves at a constant rate in the poloidal and toroidal directions, three times as fast in the poloidal direction, generating a periodic orbit. There are 100 data points in total, uniformly spaced $\pi/50$ radians apart in the toroidal direction, covering one full period of the motion. 

We use three charts (three clusters in $k$-means). The graph is built by connecting points to their two nearest neighbours. Clusters are expanded twice along the graph. 

The neural network parameters are listed in table~\ref{tab:torperSI}. The autoencoders are trained with an exponential decay learning rate schedule (initial learning rate 0.01, decay rate 0.9 every 200 steps using a staircase function), and the neural networks for dynamics are also trained with an exponential decay learning rate schedule (initial learning rate 0.01, decay rate 0.9 every 200 steps using a staircase function).

\begin{table}
\centering
\begin{tabular}{c | c | c | c} 
  \hline
  & Shape & Activations & Epochs \\
  \hline
  Encoders & 3 : 32 : 32 : 16 : 4 : 1 & elu : elu : elu : elu : linear & 1000 \\
  Decoders & 1 : 4 : 16 : 32 : 32 : 3 & elu : elu : elu : elu : linear & 1000 \\
  Dynamics & 1 : 32 : 32 : 16 : 4 : 1 & elu : elu : elu : elu : linear & 1000 \\
  \hline
\end{tabular}
\caption{Neural network parameters for periodic dynamics on a torus. ``Shape'' indicates the dimension of each layer, ``activation'' indicates the activation functions between layers, and ``epochs'' indicates the number of epochs used for training. }	
\label{tab:torperSI}
\end{table}

\subsubsection{Kuramoto-Sivashinsky beating dynamics}
\label{sec:ksbeatSI}

We compute a numerical solution to the Kuramoto-Sivashinsky equation,
\begin{equation}
  \label{eq:ksbeatSI1}
  u_t + uu_x + u_{xx} + \nu u_{xxxx} = 0,
\end{equation}
for $0 \le x \le 2\pi$ with periodic boundary conditions using a pseudo-spectral method with 64 Fourier modes. Time integration is performed using Crank-Nicolson for the linear terms and two-step Adams-Bashforth for the nonlinear term. The first step of the time integration is performed using the fourth-order Runge-Kutta method. The time step used is $10^{-4}$. The initial condition is $u(x, 0) = -\sin x + 2\cos 2x + 3\cos 3x - 4\sin 4x$ \cite{rowley2000reconstruction}. 

We set $\nu = \frac{16}{337}$; once transients decay, we obtain a beating standing wave and collect our data. The training data are the values of the field $u$ at 64 points spaced equally in $x$; our dataset consists of 100 such data points, uniformly spaced 0.01 time units apart, covering a bit more than two periods of the beating wave. 

We use three charts (three clusters in $k$-means). The graph is built by connecting points to their four nearest neighbours. Clusters are expanded once along the graph. 

The neural network parameters are listed in table~\ref{tab:ksbeatSI}. The autoencoders are trained with an exponential decay learning rate schedule (initial learning rate 0.01, decay rate 0.8 every 200 steps using a staircase function), and the neural networks for dynamics are also trained with an exponential decay learning rate schedule (initial learning rate 0.01, decay rate 0.9 every 200 steps using a staircase function). 

\begin{table}
\centering
\begin{tabular}{c | c | c | c} 
  \hline
  & Shape & Activations & Epochs \\
  \hline
  Encoders & 64 : 128 : 64 : 16 : 8 : 1 & elu : elu : elu : elu : linear & 2000 \\
  Decoders & 1 : 8 : 16 : 64: 128 : 64 & elu : elu : elu : elu : linear & 2000 \\
  Dynamics & 1 : 32 : 32 : 16 : 4 : 1 & elu : elu : elu : elu : linear & 2000 \\
  \hline
\end{tabular}
\caption{Neural network parameters for beating dynamics of the Kuramoto-Sivashinsky equation. ``Shape'' indicates the dimension of each layer, ``activation'' indicates the activation functions between layers, and ``epochs'' indicates the number of epochs used for training. }
\label{tab:ksbeatSI}
\end{table}

\subsubsection{Kuramoto-Sivashinsky beating travelling dynamics}
\label{sec:ksbtSI}

We compute a numerical solution to the Kuramoto-Sivashinsky equation as before. We set $\nu = \frac{4}{87}$; once transients decay, we obtain a beating travelling wave and collect our data. The training data are the values of the field $u$ at 64 points spaced equally in $x$; our dataset consists of 100 such data points, uniformly spaced 0.01 time units apart, covering a bit more than two beating periods and 1\% of the travelling period. Separation of the data into shape and phase variables is described in the main text. 

We use three charts (three clusters in $k$-means). The graph is built by connecting points to their four nearest neighbours. Clusters are expanded twice along the graph. 

The neural network parameters are listed in table~\ref{tab:ksbtSI}. The autoencoders are trained with an exponential decay learning rate schedule (initial learning rate 0.01, decay rate 0.9 every 200 steps using a staircase function), the neural networks for shape dynamics are also trained with an exponential decay learning rate schedule (initial learning rate 0.01, decay rate 0.9 every 200 steps using a staircase function), and the neural networks for phase dynamics are also trained with an exponential decay learning rate schedule (initial learning rate 0.01, decay rate 0.9 every 200 steps using a staircase function).

\begin{table}
\centering
\begin{tabular}{c | c | c | c} 
  \hline
  & Shape & Activations & Epochs \\
  \hline
  Encoders & 64 : 128 : 64 : 16 : 8 : 1 & elu : elu : elu : elu : linear & 2000 \\
  Decoders & 1 : 8 : 16 : 64 : 128 : 64 & elu : elu : elu : elu : linear & 2000 \\
  Shape dynamics & 1 : 32 : 32 : 16 : 4 : 1 & elu : elu : elu : elu : linear & 2000 \\
  Phase dynamics & 1 : 32 : 32 : 16 : 4 : 1 & elu : elu : elu : elu : linear & 2000 \\
  \hline
\end{tabular}
\caption{Neural network parameters for beating travelling dynamics of the Kuramoto-Sivashinsky equation. ``Shape'' indicates the dimension of each layer, ``activation'' indicates the activation functions between layers, and ``epochs'' indicates the number of epochs used for training. }
\label{tab:ksbtSI}
\end{table}

\end{document}